\documentclass[10pt,twocolumn,letterpaper]{article}

\usepackage[pagenumbers]{cvpr} 

\usepackage{mmstyle}
\usepackage{graphicx}
\usepackage{amsmath}
\usepackage{amssymb}
\usepackage{xcolor}
\definecolor{Graylight}{gray}{0.9}
\usepackage{colortbl}
\usepackage{booktabs}
\usepackage{multirow}
\usepackage{mathrsfs}
\usepackage{marvosym}
\usepackage{enumerate}
\usepackage[hang,flushmargin]{footmisc} 
\usepackage{tablefootnote}

\usepackage[accsupp]{axessibility} 

\usepackage[pagebackref,breaklinks,colorlinks]{hyperref}

\def\paperID{} 
\def\confName{CVPR}
\def\confYear{2024}

\begin{document}


\title{InternLM-XComposer: A Vision-Language Large Model for \\ Advanced Text-image Comprehension and Composition}

\author{Pan Zhang$^{*1}$, Xiaoyi Dong$^{*1}$, Bin Wang$^{1}$, Yuhang Cao$^{1}$, Chao Xu$^{1}$, Linke Ouyang$^{1}$, Zhiyuan Zhao$^{1}$, \\ Haodong Duan$^{1}$, Songyang Zhang$^{1}$, Shuangrui Ding$^{1}$, Wenwei Zhang$^{1}$, Hang Yan$^{1}$, Xinyue Zhang$^{1}$, Wei Li$^{1}$, \\ Jingwen Li$^{1}$,
Kai Chen$^{1}$, Conghui He$^{1}$, Xingcheng Zhang$^{1}$, Yu Qiao$^{1}$, Dahua Lin$^{1}$, Jiaqi Wang$^{1,}${\textsuperscript{\Letter}}\\
$^1$Shanghai Artificial Intelligence Laboratory \\
\tt\small
internlm@pjlab.org.cn
}

\twocolumn[{
\renewcommand\twocolumn[1][]{#1}
\maketitle
\begin{center}
    \centering
    \vspace{-20pt}
    \includegraphics[width=1.0\linewidth]{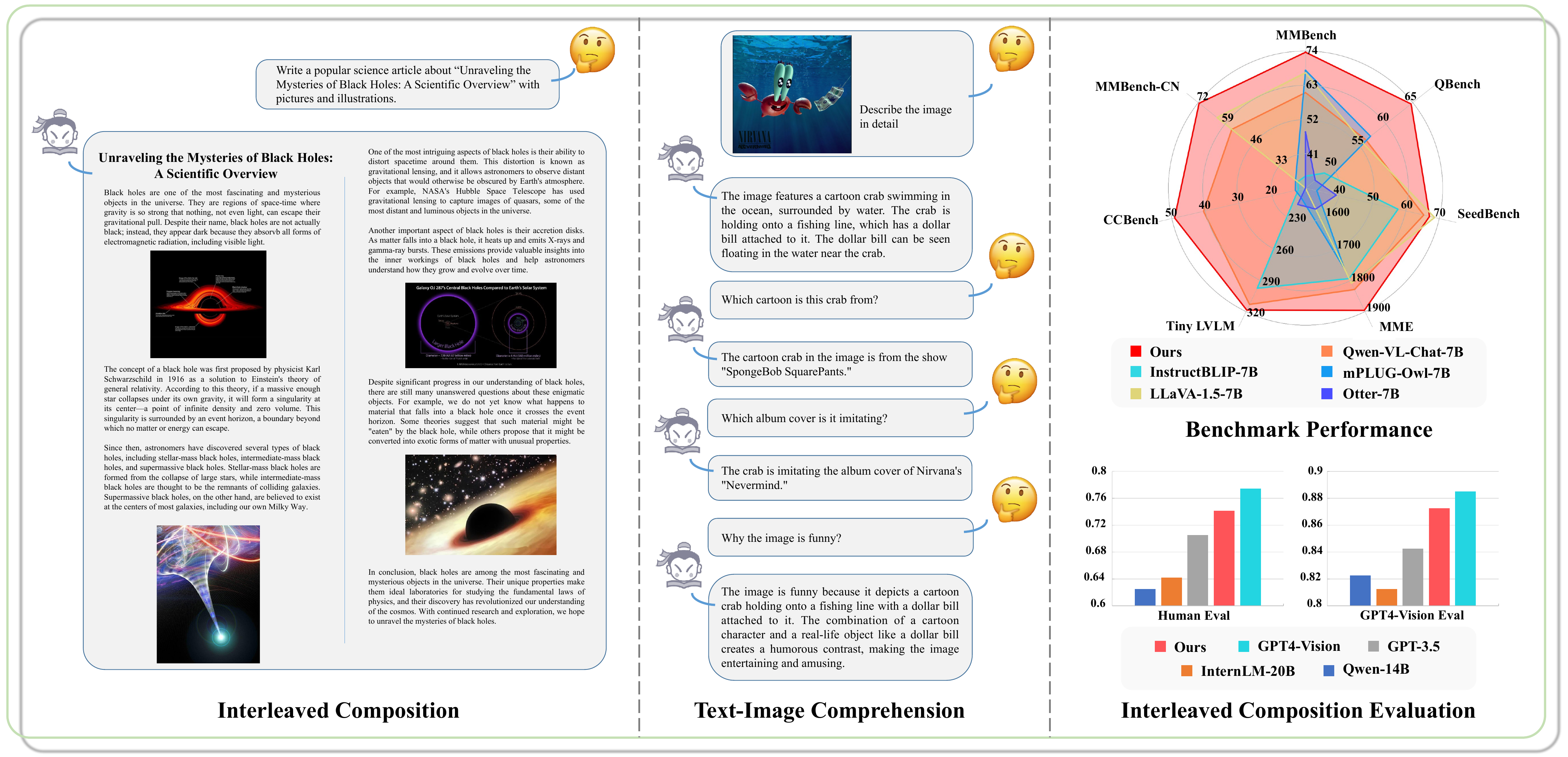}
    \setlength{\abovecaptionskip}{0mm}
    \vspace{-10pt}
    \captionof{figure}{\small
        The InternLM-XComposer shows excellent interleaved composition and text-image comprehension ability, leading to strong performance on various multi-modal benchmarks. For the Interleaved Composition Evaluation by both Human and GPT4-Vision (GPT4-V), our model shows competitive performance to GPT4-Vision and GPT3.5.
	}
	\label{fig:teaser}
    \vspace{-5pt}
\end{center}
}]

\maketitle

{\let\thefootnote\relax\footnotetext{\noindent* indicates equal contribution.}}

\begin{abstract}

We propose InternLM-XComposer, a vision-language large model that enables advanced image-text comprehension and composition. The innovative nature of our model is highlighted by three appealing properties:
1) \textbf{Interleaved Text-Image Composition}: 
InternLM-XComposer can effortlessly generate coherent and contextual articles that seamlessly integrate images, providing a more engaging and immersive reading experience. Simply provide a writing instruction, and our system will generate the corresponding manuscript. It can intelligently identify the areas in the text where images would enhance the content and automatically insert the most appropriate visual candidates.
2) \textbf{Comprehension with Rich Multilingual Knowledge}: The text-image comprehension is empowered by training on an extensive multi-modal multilingual database with carefully crafted strategies, resulting in a deep understanding of visual content.
3) \textbf{State-of-the-art Performance}: Our model consistently achieves state-of-the-art results across various mainstream benchmarks for vision-language foundational models, including MME Benchmark, MMBench, MMBench-CN, Seed-Bench, CCBench (Chinese Cultural Benchmark), QBench and Tiny LVLM.
Owing to the absence of established metrics for quantitatively assessing text-image composition, we have devised a robust evaluation procedure that comprises both human and GPT4-Vision (GPT4-V) to ensure reliability. Notably, our InternLM-XComposer achieves competitive text-image composition scores compared to public solutions, including GPT4-V and GPT3.5.
Collectively, InternLM-XComposer seamlessly blends advanced text-image comprehension and composition, revolutionizing vision-language interaction and offering new insights and opportunities. 
The InternLM-XComposer model series are publicly available at \url{https://github.com/InternLM/InternLM-XComposer}.
\end{abstract}

\section{Introduction}
\label{sec:intro}
Over the past year, impressive progress has been made in developing large language models (LLMs)~\cite{raffel2020exploring,brown2020language,chowdhery2022palm,openai2020chatgpt,openai2023gpt4,touvron2023llama,vicuna2023,qwen7b,touvron2023llama2,baichuan2023baichuan2,2023internlm,du2022glm}. These state-of-the-art models, including ChatGPT~\cite{openai2020chatgpt}, GPT4~\cite{openai2023gpt4}, and PaLM 2~\cite{chowdhery2022palm}, have shown an unprecedented ability to follow human instructions and solve open-ended tasks. 
Inspired by the success of PaLM-E~\cite{driess2023palme} and BLIP2~\cite{Li2023BLIP2BL}, there is a promising approach to extending language models for vision-language tasks by leveraging vision features as extra inputs of LLMs. The community has developed several vision-language large models (VLLMs), such as MiniGPT-4~\cite{zhu2023minigpt}, LLaVA~\cite{liu2023visual}, and InstructBLIP~\cite{dai2023instructblip}, based on open-source LLMs like LLaMA~\cite{touvron2023llama}, GLM~\cite{du2022glm}, and InternLM~\cite{2023internlm}.
However, these VLLMs focus on pure text outputs, missing the opportunity to equip generated text with richer information through auxiliary multimedia content like images.

In this work, we propose InternLM-XComposer, which is a vision-language large model that enables advanced text-image comprehension and composition ability.

1) \textbf{Interleaved Text-Image Composition}.
InternLM-XComposer excels in generating long-form content that is interleaved with contextually relevant images, providing more engaging and immersive vision-language interactions. In its operational flow, the framework first crafts articles following human-provided instructions. Subsequently, it autonomously pinpoints optimal locations within the text for image placement and furnishes corresponding suitable image descriptions. 
In accordance with the generated descriptions, instead of relying on text-to-image generation models~\cite{DALLE3,DALLE2,DALLE,Rombach_2022_CVPR,Imagen}, we opt to source aligned images from a large-scale web-crawled image database for realistic quality and contextual alignment. Moreover, it also provides flexibility by allowing users to customize an image repository.

Compared to a baseline approach that relies solely on CLIP~\cite{radford2021learning,Yang2022ChineseCC} for image retrieval, XComposer offers a more reliable solution for choosing the most appropriate image. Initially, we select potential image candidates from our database using CLIP. Then, InternLM-XComposer leverages its comprehension capabilities to identify the image that optimally complements the content.
   
2) \textbf{Comprehension with Rich Multilingual Knowledge}. 
LLM demonstrates remarkable ability in handling open-world tasks, a capability attributed to its extensive training data, \eg, 2T text tokens used in LLaMA2~\cite{touvron2023llama2}. This vast dataset inherently encapsulates a broad spectrum of semantic knowledge across diverse domains. In contrast, current vision-language datasets~\cite{sharma2018conceptual,Ordonez_2011_im2text,schuhmann2021laion,changpinyo2021conceptual} are limited in both volume and diversity compared to the LLM database. This results in insufficient coverage of vision-language concepts due to the long-tailed data distribution~\cite{Newman_2005,Liu_2019,cui_2019} in the real world. To tackle this limitation, we employ two practical solutions: First, to extend the knowledge of VLLMs with widespread concepts, a multilingual vision-language dataset comprising over 11 million semantic concepts is collected from public websites. We further gather open-sourced vision-language datasets to organize our high-quality training data. Second, we carefully crave the pretraining and finetuning strategies in our training pipeline, where we adopt the mixed training data of pure text and image-text data, primarily in English and Chinese, to keep the initial capabilities of LLMs.
Consequently, InternLM-XComposer demonstrates a remarkable proficiency in comprehending image content and responding with extensive multilingual knowledge. 
InternLM-XComposer stands out for its unique ability to compose long-form articles that incorporate contextually relevant images. This process involves creating high-quality written content, identifying appropriate positions for inserted images, and selecting the most suitable images that consider the complex context of interleaved text and images.

The proposed InternLM-XComposer exhibits superior capabilities in both text-image comprehension and composition, as evidenced by its strong performance in quantitative benchmarks and compelling qualitative demonstrations. It consistently achieves \textbf{state-of-the-art} performance across various leading benchmarks for vision-language large models, encompassing MME Benchmark~\cite{yin2023survey,fu2023mme}, MMBench~\cite{MMBench}, Seed-Bench~\cite{li2023seedbench}, QBench~\cite{wu2023q}, Tiny LVLM~\cite{shao2023tiny} in English, and MMBench-CN~\cite{MMBench}, CCBench (Chinese Cultural Benchmark)~\cite{MMBench} for evaluations in Chinese.
Furthermore, in response to the absence of established metrics for quantitatively assessing text-image composition, we set up an evaluation procedure for interleaved text-image articles. This procedure takes into account the quality of both the written content and the accompanying illustrated images, which harnesses both human assessment and GPT4-Vision (GPT4-V) scoring to enhance robustness and reliability. The evaluation outcomes, spanning assessments by both human and GPT4-V, consistently demonstrate the competitive performance of InternLM-XComposer in interleaved text-image composition compared to public solutions, including GPT3.5 and GPT4-V.

\section{Related Works}
\label{sec:related}

\noindent{\textbf{Large Language Models (LLMs).}}
In recent years, the development of large language models has accelerated. Initially, encoder-decoder models such as BERT~\cite{devlin2018bert} and T5~\cite{raffel2020exploring}, as well as decoder-only models like GPT~\cite{radford2018improving}, leveraged the Transformer architecture~\cite{vaswani2017attention} to achieve remarkable results. GPT3~\cite{brown2020language}, employing prompt and in-context learning strategies along with larger models and data, has significantly performed in few-shot and zero-shot tasks. As a result, using decoder-only structures with autoregressive training has gained popularity among researchers. Google's PaLM~\cite{chowdhery2022palm} further expands the model parameter size and data volume. To enhance the conversational experience, models like InstructGPT~\cite{ouyang2022training} and ChatGPT~\cite{openai2020chatgpt} integrate instruction-tuning and reinforcement learning from human feedback (RLHF). The open-sourcing of LLaMA~\cite{touvron2023llama} model has inspired research on LLMs, \eg, Alpaca~\cite{alpaca}, Vicuna~\cite{vicuna2023}, GLM~\cite{du2022glm,zeng2023glm-130b}, Qwen~\cite{qwen7b}, LLaMA2~\cite{touvron2023llama2}, Baichuan2~\cite{baichuan2023baichuan2},  InternLM~\cite{2023internlm}, Falcon~\cite{penedo2023refinedweb}, and Mistral~\cite{jiang2023mistral}.

\noindent{\textbf{Vision Large Language Models (VLLMs).}}
Visual language learning has emerged as a research hotspot. CLIP~\cite{radford2021learning} and its following works~\cite{li2022blip,li2021grounded,zhang2022glipv2,liu2023grounding} aligns image and text features through contrastive learning objectives on large-scale image-text pairs, outperforming supervised learning on ImageNet~\cite{deng2009imagenet} and exhibiting strong generalization capabilities in various downstream tasks.
However, these models show limited capabilities for tasks requiring higher-level understanding, such as visual question answering. Benefiting from existing large language model~\cite{raffel2020exploring,touvron2023llama,vicuna2023} and visual encoder~\cite{li2022blip,fang2023eva}, the vision large language models (VLLMs)~\cite{chen2023pali,chen2023palix,chen2023pali3,driess2023palme} achieve fine-grained alignment between the visual information and the LLMs, show superb performance in diverse tasks. 
To achieve the modality alignment, a series of studies\cite{liu2023visual,wang2023vigc,zhao2023mllm,zhao2023mmicl,li2023otter,chen2023shikra,peng2023kosmos,ye2023mplug,awadalla2023openflamingo,alayrac2022flamingo,chen2023minigptv2} have explored the impact of the quality, diversity, and specificity of the fine-tuning data and the learnable parameters. For example, MiniGPT4~\cite{zhu2023minigpt} adopts a simple FC layer with a small amount of caption data. InstructBLIP~\cite{dai2023instructblip} fine-tunes the Q-Former~\cite{li2022blip} on diverse image-text tasks. LLaVA~\cite{liu2023visual} fine-tunes the LLM with GPT-4 generated high-quality instruction tuning data. Qwen-VL~\cite{bai2023qwen} and CogVLM~\cite{wang2023cogvlm} fine-tune on high-resolution images with multi-task training.
Moreover, some recent works~\cite{emu_2023,seed_2023,CM3Leon,dong2023dreamllm} integrate the image generation task with VLLMs to generate text-aligned images. InternLM-XComposer stands out for its unique ability to compose long-form articles that incorporate contextually relevant images. This process involves creating high-quality written content, identifying appropriate positions for inserted images, and selecting the most suitable visual content that considers the interleaved vision-language context.

\noindent{\textbf{Image-text Retrieval Models.}}
Image-text retrieval, a pivotal area in multimodal modeling, has seen substantial advancements recently. CLIP~\cite{radford2021learning} and its following works~\cite{jia2021scaling,hu2023reveal,li2022blip,yasunaga2023retrieval,koh2023grounding,zhang2205opt} utilize contrastive learning on a large corpus of image-text pairs, excels in image-text matching, enabling efficient retrieval in both image-to-text and text-to-image modalities. However, the capabilities of current models are primarily confined to matching images with aligned descriptions, which has a significant gap between our image-text article composition task, which focuses on selecting a suitable image based on a complex context containing interleaved images and long text.

\section{Method}

\subsection{Model Architecture}
\begin{figure}[t]
	\centering
	\includegraphics[width=1\linewidth,]{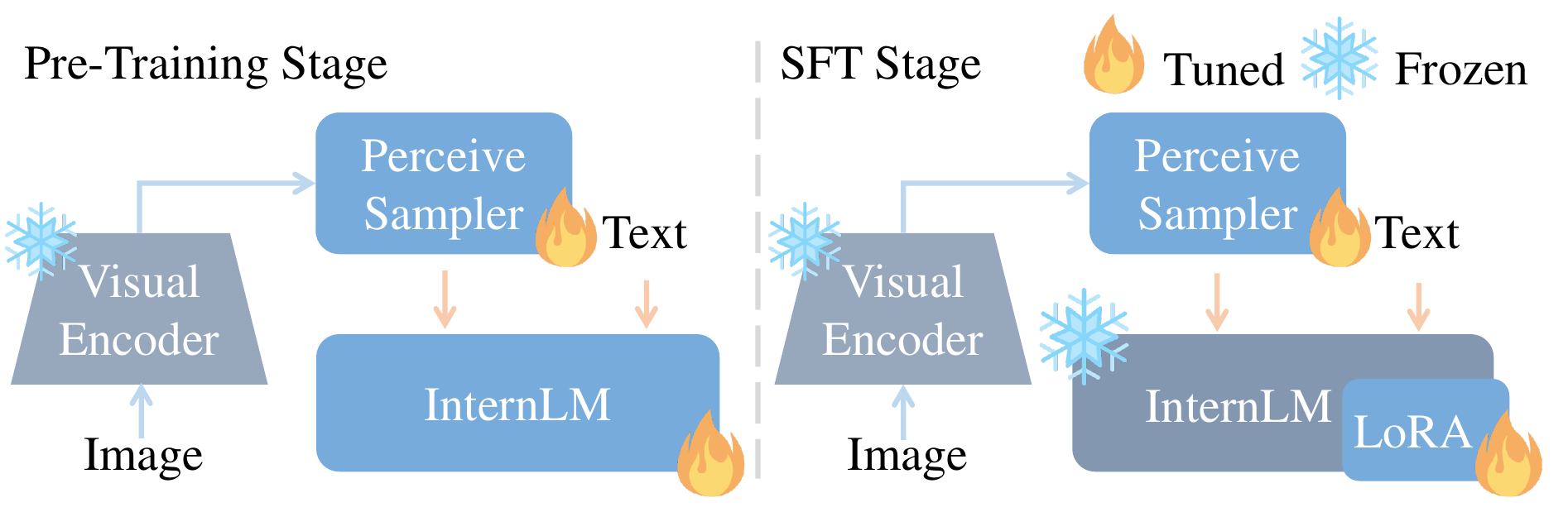}
	\caption{The architecture and training regimen of the InternLM-XComposer. The Pre-training stage aligns the visual and language knowledge and the SFT stage excites different model capabilities.
 }
	\label{fig:architecture}
 \vspace{-5pt}
\end{figure}

As depicted in Figure~\ref{fig:architecture}, the proposed InternLM-XComposer contains three integral components: a visual encoder, a perceive sampler, and a large language model.

\noindent{\textbf{Visual Encoder.}} The visual encoder in InternLM-XComposer employs EVA-CLIP~\cite{fang2023eva}, an refined variant of the standard CLIP~\cite{Yang2022ChineseCC}, enhanced with mask image modeling capabilities, to proficiently capture the visual nuances of the input image. Within this module, images are resized to a consistent dimension of $224\times 224$ and subsequently dissected into patches with a stride of 14. These patches serve as input tokens and enable the self-attention mechanisms within the transformer block, facilitating the extraction of detailed image embeddings.

\begin{table}[t]
\centering
\small
\setlength{\tabcolsep}{1.2mm}{
\begin{tabular}{ccccc}
\toprule
Lauguage & Type & Dataset &  Images &  Text\\
\midrule
 \multirow{5}{*}{English} 
 & Pure Text & WanJuan~\cite{He2023WanJuanAC} & - & 5B \\
 & Paired & CC 3M~\cite{sharma2018conceptual} & 3M & 37M \\
 & Paired & SBU-Caption~\cite{Ordonez_2011_im2text}) & 1M & 18M \\
 & Paired & LAION400M~\cite{schuhmann2021laion} & 509M & 10B \\
 & Paired & CC 12M~\cite{changpinyo2021conceptual} & 9M & 250M \\
& Paired & In-house Concept data & 2M & 321M \\
 & Interleaved   & Multimodal C4~\cite{zhu2023multimodal} & 332M & 40B \\
\midrule
\multirow{5}{*}{Chinese}
& Pure  Text & WanJuan~\cite{He2023WanJuanAC} & - & 5B \\
& Paired & TaiSu~\cite{liu2022taisu} & 44M & 865M \\
& Paired & WuKong~\cite{gu2022wukong} & 31M & 545M \\
& Paired & LAION-CN~\cite{schuhmann2022laion} & 80M & 2B \\
& Paired & In-house Concept data & 9M & 704M \\
& Interleaved  & WanJuan~\cite{He2023WanJuanAC} & 85M & 13B \\
\midrule
&  & Total & 1.1B & 77.7B \\
\bottomrule
\end{tabular}}
\caption {Details of InternLM-XComposer pre-training data. LAION-CN represents the Chinese language subset extracted from the larger LAION-5B corpus. This subset is further cleaned utilizing the Chinese CLIP~\cite{Yang2022ChineseCC}. The volume of text data is counted in terms of the number of tokens. The In-house Concept data is collected from public websites, including over 11 million vision-language concepts from public websites.}
\vspace{-5pt}
\label{tab:pre-training data}
\end{table}

\noindent{\textbf{Perceive Sampler.}} The perceive sampler within the InternLM-XComposer operates as an attentive pooling mechanism designed to condense the initial set of 257 image embeddings down to 64 refined embeddings. These optimized embeddings are subsequently aligned to be compatible with the knowledge structures understood by the large language model. Following BLIP2~\cite{Li2023BLIP2BL}, we leverage $\text{BERT}_{base}$~\cite{devlin2018bert} equipped with cross-attention layers, serving as the perceive sampler in our framework.

\noindent{\textbf{Large Language Model.}} The InternLM-XComposer is anchored on InternLM~\cite{2023internlm} as its foundational large language model. Notably, InternLM stands as a potent language model equipped with multilingual capabilities, proficient in both English and Chinese. In our framework, we employ the publicly available InternLM-Chat-7B to serve as the large language model.

\begin{table}[t]
\centering
\footnotesize
\setlength{\tabcolsep}{1.5mm}{
\begin{tabular}{ll}
\toprule
Task &  Dataset\\
\midrule
\multicolumn{2}{l}{\textit{Multi-task training}} \\
Caption  &  COCO~\cite{chen2015microsoft}, SUB~\cite{chen2015microsoft}, TextCaps~\cite{sidorov2020textcaps} \\
VQA      &  VQAv2~\cite{VQAv2}, GQA~\cite{hudson2018gqa}, OK-VQA~\cite{marino2019ok},\\
& IConQA~\cite{lu2021iconqa}  Text-VQA~\cite{singh2019towards}, SQA~\cite{lu2022learn}, \\
&  VSR~\cite{Liu2022VisualSR}, OCR-VQA~\cite{mishraICDAR19},  VIGC~\cite{wang2023vigc}\\
IQG          & VQAv2~\cite{VQAv2}, OK-VQA~\cite{marino2019ok}, A-OKVQA~\cite{schwenk2022okvqa} \\
Conversation  &  Visual Dialog~\cite{das2017visual}, LLaVA-150k~\cite{liu2023visual} \\
\midrule
\multicolumn{2}{l}{\textit{Instruction tuning}} \\
Composiiton  &  In-house data (Refer to Sec.\ref{sec:compos_data})\\
Conversation  &  LLaVA-150k~\cite{liu2023visual}, Alpaca-en\&zh~\cite{alpaca}\\
& ShareGPT-en\&zh ~\cite{vicuna2023}, Oasst-en\&zh ~\cite{oasst}, LRV~\cite{liu2023aligning} \\
\bottomrule

\end{tabular}}
\vspace{-5pt}
\caption {Datasets used for Supervised Fine-Tuning. }
\vspace{-15pt}
\label{tab:sft data}
\end{table}

\subsection{Training}
As shown in Figure~\ref{fig:architecture}, the training process of InternLM-XComposer is split into Pre-training Stage and Supervied Fine-tuning (SFT) Stage. The pre-training stage utilizing vast amounts of data for foundation model training, aligning the knowledge between the visual information and language. 
Based on the pre-training model, the SFT stage involves a multi-task training step and a following instruction tuning step to excite different model capabilities. 

\noindent{\textbf{Pre-training.}} The pre-training phase incorporates large-scale image-text pairs along with interleaved image-text data to pre-train the foundational vision-language model. This data comprises multi-modal content in both English and Chinese. As shown in Table~\ref{tab:pre-training data}, in addition to public datasets, to enhance the ability of MLLMs to comprehend extensive visual concepts, we collect large-scale \textbf{In-house Concept data}\footnote{Due to copyright concerns, we will release the concept name list and source websites for the community to reproduce our concept data.} from various public websites, including more than 11 million visual concepts and corresponding details explanations. Please refer to supplementary materials for more information. Moreover, to preserve the capabilities of the initial large language model, the partial textual data~\cite{He2023WanJuanAC} utilized for InternLM's pre-training is also employed in the pre-training phase of InternLM-XComposer. The multi-modal pre-training process employs 1.1 billion images alongside 77.7 billion text tokens as in Table~\ref{tab:pre-training data}.

During the pre-training phase, the visual encoder is frozen, allowing the optimization to be concentrated on the perceive sampler and the large language model. Initial weight for the perceive sampler and the large language model are sourced from BLIP2~\cite{Li2023BLIP2BL} and InternLM~\cite{2023internlm}, respectively. Given that the large language model lacks native understanding of image embeddings, its optimization within the framework of multimodal pre-training serves to enhance its capability to interpret such embeddings effectively. The training objective for the model centers on next-token prediction, utilizing cross-entropy loss function.

\noindent{\textbf{Supervised Fine-tuning.}}
In the pre-training phase, image embeddings are aligned with language representations, equipping the large language model with a rudimentary understanding of image content. However, the model still lacks proficiency in utilizing the image information optimally. To address this limitation, we introduce a variety of vision-language tasks that the model undertakes during the subsequent Supervised Fine-Tuning Stage, which contains two consecutively steps, \ie, \emph{Multi-task training} and \emph{Instruction tuning}.

\begin{figure*}[t!]
	\centering
	\includegraphics[width=0.95\linewidth]{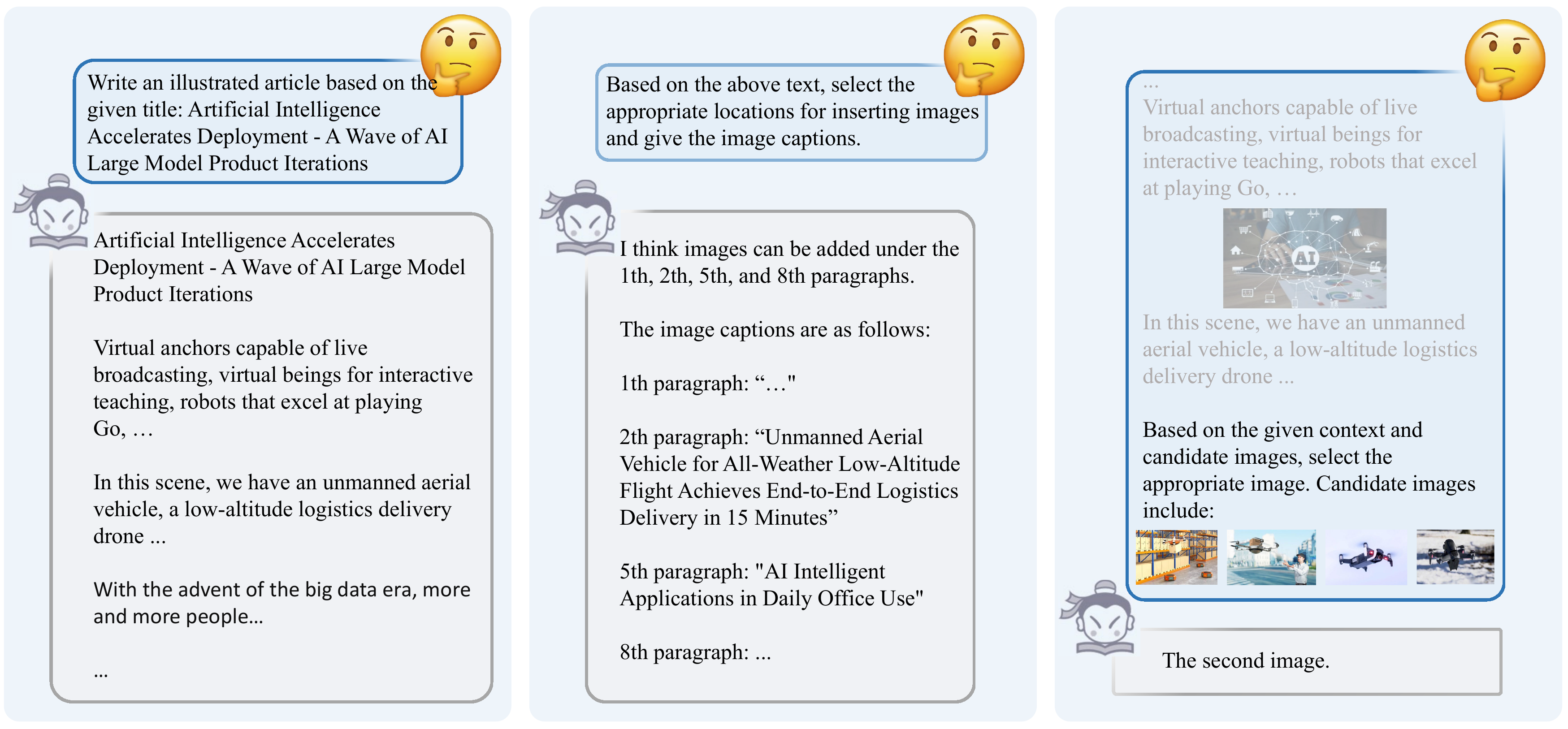}
        \put(-420,-8){\footnotesize (a) Text Generation}
        \put(-295,-8){\footnotesize (b) Image Spotting and Captioning}
        \put(-135,-8){\footnotesize (c) Image Retrieval and Selection}
	\caption{\textbf{The pipeline of the interleaved image-text composition.} (a) Given a writing instruction, the model initially generates a corresponding text-based article. (b) Subsequent to the article generation, the model is trained to identify suitable image locations and generate corresponding captions for the ensuing steps. (c) A text-image retrieval algorithm is initially employed to constrict the pool of candidate images. Following this, our model leverages its vision-language understanding ability to make the final image selection, ensuring thematic and visual coherence by considering both the preceding textual content and images within the article.}
	\label{fig:interleaved}
 \vspace{-10pt}
\end{figure*}

\noindent\emph{Multi-task Training}. As illustrated in Table~\ref{tab:sft data}, the multi-task training dataset is constructed from multiple sources to endow the model with a diverse range of capabilities, including scene understanding (\eg, COCO Caption~\cite{chen2015microsoft}, SUB~\cite{Ordonez_2011_im2text}), location understanding (\eg, Visual Spatial Reasoning dataset~\cite{Liu2022VisualSR}), optical character recognition (OCR) (\eg, OCR-VQA~\cite{mishraICDAR19}), and open-ended answering (\eg, VQAv2~\cite{VQAv2}, GQA~\cite{hudson2018gqa}), among others. Each of these tasks is formulated as a conversational interaction, adhering to the following format:
\begin{align*} 
& \texttt{<|User|>}: \textit{Instruction} \enspace \texttt{<eou>} \\
& \texttt{<|Bot|>}: \textit{Answer} \enspace \texttt{<eob>} 
\end{align*}
where $\texttt{<eou>}$ and $\texttt{<eob>}$ represent the \textit{end-of-user} and \textit{end-of-bot} tokens, respectively. For VQA datasets with multiple questions per image, we structure them as multi-round conversations with randomly ordered questions, thereby substantially enhancing the efficiency of SFT.

In order to achieve stable and efficient fine-tuning, we retains the weights of the pre-existing large language model in a frozen state. Subsequently, we augment the architecture with Low-Rank Adaption (LoRA)~\cite  {hu2022lora} for the fine-tuning process. The perceive sampler is concurrently trained, albeit with a distinct learning rate. 

\noindent\emph{Instruction Tuning}. To further empower aforementioned model's instruction following and interleaved image-text composition capability, as shown in Table~\ref{tab:sft data}, we utilize data from pure-text conversation corpora and LLaVA-150k for instruction-based tuning, and leverage the LRV dataset to mitigate hallucinations. The interleaved image-text composition dataset is constructed based on the methodology delineated in Section \ref{sec:compos_data}.

\subsection{Interleaved Image-Text Composition}
\label{sec:compos_data}
To craft interleaved image-text compositions, the initial step involves the generation of an article following human instruction. After this, accompanying illustrated images are incorporated at well-suited positions within the textual content, enriching the overall narrative and augmenting reader engagement. The pipeline of interleaved image-text composition is shown in Figure~\ref{fig:interleaved}.

\noindent{\textbf{Text Generation.}} To facilitate the generation of extended text-based articles, we collect a dataset comprising interleaved image-text compositions from the Internet.
To enable the model to generate text-based articles with respect to specific instructions, we utilize GPT-4 to generate diverse instructions based on the article, including a draft outline, a few keywords, or a simple title.

As mentioned in Section~\ref{sec:intro}, a more engaging and immersive reading experience needs both text and suitable images. We empower our model with such composition capability with a decoupled pipeline, which defines the position and caption of images based on the generated text, and then selects the one image for each position from a list of candidates according to article context.

\noindent{\textbf{Image Spotting and Captioning.}} 
First, the acquired interleaved image-text data is leveraged to train our model for pinpointing image locations. For subsequent image retrieval, it's imperative to generate an appropriate caption, enabling the application of various text-image retrieval algorithms. 
To mitigate this challenge, we generate caption data utilizing GPT-4, which is provided with the text-based article and image locations and is instructed to generate a caption for each image position that remains coherent with the overarching theme and concept.

\begin{table*}[t]
\small
\centering
\setlength{\tabcolsep}{0.7mm}{
\begin{tabular}{lcccccccccccccccc}
\toprule

Model & Overall & Exist. & Count & Pos. & Color & OCR & Poster & Cele. & Scene & Land. & Art. & Comm. & NumCal. & Trans. & Code & Avg. \\
\hline

MiniGPT-4\cite{zhu2023minigpt} & 694.3 & 68.3 & 55.0 & 43.3 & 43.3 & 57.5 & 41.8 & 54.4 & 71.8 & 54.0 & 60.5 & 59.3 & 45.0 & 0.0 & 40.0  & 49.6 \\

LLaVA\cite{liu2023visual} & 712.5 & 50.0 & 50.0 & 50.0 & 50.0 & 50.0 & 50.0 & 48.8 & 50.0 & 50.0 & 49.0 & 57.1 & 50.0 & 57.5 & 50.0 & 50.9  \\

VisualGLM\cite{du2022glm} & 880.4 & 85.0 & 50.0 & 48.3 & 48.3 & 42.5 & 66.0 & 53.2 & 146.3 & 83.8 & 75.3 & 39.3 & 45.0 & 50.0 & 47.5 & 62.9  \\

mPLUG-Owl\cite{ye2023mplug} & 1238.4 & 120.0 & 50.0 & 50.0 & 50.0 & 65.0 & 136.1 & 100.3 & 135.5 & 159.3 & 96.3 & 78.6 & 60.0 & 80.0 & 57.5 & 88.5 \\

LLaMA-A.-V2\cite{Gao2023LLaMAAdapterVP} & 1194.9 & 120.0 & 50.0 & 48.3 & 48.3 & \underline{125.0} & 99.7 & 86.2 & 148.5 & 150.3 & 69.8 & 81.4 & 62.5 & 50.0 & 55.0 & 85.4 \\

InstructBLIP\cite{dai2023instructblip} & 1417.9 & 185.0 & 143.3 & 66.7 & 66.7 & 72.5 & 123.8 & 101.2 & 153.0 & 79.8 & \underline{134.3} & 129.3 & 40.0 & 65.0 & 57.5 & 101.3\\

Lynx\cite{Zeng2023WhatMI} & 1508.9 & \underline{195.0} & {151.7} & 90.0 & 90.0 & 77.5 & 124.8 & 118.2 & \underline{164.5} & {162.0} & 119.5 & 110.7 & 17.5 & 42.5 & 45.0 &  107.8\\

Otter\cite{li2023otter} & 1572.0 & \underline{195.0} & 88.3& 86.7 & 86.7 & 72.5 & 138.8 & \underline{172.7} & {158.8} & 137.3 & \underline{129.0} & 106.4 & \underline{72.5} & 57.5 & 70.0 & 112.3 \\

Cheetor\cite{Li2023EmpoweringVM} & 1584.4 & 180.0 & 96.7 & 80.0 & 80.0 & 100.0 & 147.3 & \underline{164.1} & 156.0 & 145.7 & 113.5 &98.6 & \underline{77.5} & 57.5 & \underline{87.5} & 113.2 \\

BLIVA\cite{Hu2023BLIVAAS}  & 1669.2 & 180.0 & 138.3 & 81.7 & \underline{180.0} & 87.5& {155.1} & 140.9 & 151.5 & 89.5 & \underline{133.3}  & \underline{136.4} & 57.5 & 77.5 & 60.0 & 119.2 \\

MMICL\cite{Zhao2023MMICLEV} & 1810.7 &170.0 & \underline{160.0} & 81.7 & 156.7  & 100 & 146.3 & 141.8 & 153.8 & 136.1 & \underline{135.5} & \underline{136.4} & \underline{82.5}	&\underline{132.5} & \underline{77.5} & 129.3 \\

LLaVA-1.5\cite{llava1_5} & 1826.7 & 185.0 & \underline{155.0} & \underline{133.3} & \underline{170.0} & 125.0 & \underline{160.5} & \underline{152.9} & \underline{161.3} & \underline{170.5} & 117.7 & 127.8 & 42.5 & 77.5 & 47.5 & 130.5 \\

Qwen-VL-Chat\cite{bai2023qwen} & 1848.3 & 158.3 & 150.0 & \underline{128.3} & \underline{170.0} & \underline{140.0}& \underline{178.6} & 120.6 & 152.3 & \underline{164.0} & 125.5  & 130.7 & 40.0 & \underline{147.5} & 42.5 & 132.0 \\

\midrule

\rowcolor{Graylight} Ours & \textbf{1919.5} & \underline{190.0} & \underline{158.3} & \underline{126.7} & {165.0} & \underline{125.0} & \underline{161.9} & {150.3} & \underline{159.8} & \underline{165.3} & 126.3 & \underline{138.6} & 55.0 & \underline{112.5} & \underline{85.0} & \textbf{137.1}  \\

\bottomrule
\end{tabular}%
}
\vspace{-3mm}
\caption{
\textbf{Evaluation of MME-Benchmark}. Here we report the results on all the sub tasks, including Existence(Exist.), Count, Position(Pos.), Color, OCR, Poster, Celebrity(Cele.), Scene, Landmark(Land.), Artwork(Art.), Commonsense Reasoning(Comm.), Numerical Calculation(NumCal.), Text Translation(Trans.), Code Reasoning(Code) and the task-level average (Avg.). We \textbf{bold} the \emph{highest average / overall score} and highlight the \emph{Top-3} model of each \emph{sub task} with \underline{underline}. }
\vspace{-10 pt}
\label{table:mme}
\end{table*}

\noindent{\textbf{Image Retrieval and Selection.}}
Having obtained the captions, a variety of text-image retrieval methods become available for use. In this work, we opt for the CLIP model, capitalizing on its proven efficacy. We compute the similarity scores between the generated caption and each image in the candidate pool. The top \( m \) images, based on these similarity scores, are then selected to constitute the reduced candidate pool for further processing.

To guarantee thematic or conceptual coherence in the images dispersed throughout the article, we deploy our vision-language model to execute the final image selection. Our model selects images by considering both preceding text and prior images within the article. This mechanism enables the model to acquire an understanding of thematic and visual coherence, an expertise derived from the dataset of interleaved image-text compositions. Our model is also trained to select the ground-truth image from a candidate list given the article and image position to enhance the capability. The training data is directly structured from collected interleaved text-image articles and randomly selected negative candidates from the image repository.
\section{Experiments}

\subsection{English-Based Benchmark results.}
In this section, we validate the benchmark performance of our InternLM-XComposer after the Multi-task training. In the following, the comparison model is 7B version by default if there is no additional annotation.

\begin{table}[!t]
\small
\centering
\setlength{\tabcolsep}{1.mm}{
\begin{tabular}{lcccccccccc}
\toprule
Method &  Avg. & LR & AR & RR & FP-S & FP-C & CP \\
\midrule 
MiniGPT-4\cite{zhu2023minigpt} &  23.0 & 13.6 & 32.9 & 8.9 & 28.8 & 11.2 & 28.3\\ 
VisualGLM \cite{du2022glm}&  33.5 & 11.4 & 48.8 & 27.7 & 35.8 & 17.6 & 41.5 \\
InstructBLIP\cite{dai2023instructblip}  & 33.9 & 21.6 & 47.4 & 22.5 & 33.0 & 24.4 & 41.1 \\
LLaVA \cite{liu2023visual}&  36.2 & 15.9 & 53.6 & 28.6 & 41.8 & 20.0 & 40.4 \\
LLaMA-A-V2 \cite{Gao2023LLaMAAdapterVP}&  38.9 & 7.4 & 45.3 & 19.2 & 45.0 & 32.0 & 54.0 \\
Otter-I \cite{li2023otter}&  48.3 & 22.2 & 63.3 & 39.4 & 46.8 & 36.4 & 60.6 \\
LLaVA-1.5 \cite{llava1_5}& 59.5 & 32.4 & 72.6 & 49.3 & 62.3 & 52.2 & 67.7 \\
Shikra \cite{chen2023shikra}&  60.2 & 33.5 & 69.6 & 53.1 & 61.8 & 50.4 & 71.7 \\
Qwen-VL-Chat \cite{bai2023qwen}& 61.8 & 40.5 & 74.3 & 47.9 & 66.3 & 46.2 & 72.8 \\
LMEye \cite{li2023lmeye}& 62.6 & 41.0 & 74.3 & 55.9 & 61.6 & 58.7 & 69.2 \\
MMICL\cite{zhao2023mmicl} &  65.2 & 44.3 & 77.9 & 64.8 & 66.5 & 53.6 & 70.6 \\ 
mPLUG-Owl\cite{ye2023mplug} & 66.0 & 43.4 & 76.0 & 62.1 & 68.6 & 55.9 & 73.0 \\

\midrule 
\rowcolor{gray!20}Ours &  \textbf{74.4} & \textbf{50.6} & \textbf{82.0} & \textbf{76.1} & \textbf{79.3} & \textbf{59.2} & \textbf{81.7} \\
\bottomrule
\end{tabular}%
}
\vspace{-3mm}
\caption{
\textbf{Evaluation of MMBench test set.} Here we report the results on the six L-2 abilities, namely Logical Reasoning (LR), Attribute Reasoning (AR), Relation Reasoning (RR), Fine-grained Perception (Cross Instance) (FP-C), Fine-grained Perception (Single Instance) (FP-S), and Coarse Perception (CP).
}
\vspace{-15pt}
\label{table:mmbench_results}
\end{table}

\noindent\textbf{MME Benchmark}\cite{fu2023mme} measures the perception and cognition capability of multi-modality LLMs with carefully crafted questions within 14 sub-tasks. As shown in Table.\ref{table:mme}, our model reached a new state-of-the-art performance $137.11\%$, outperforms the previous method Qwen-VL-Chat with more than $5.0\%$. We also highlight the Top-3 models within each subtask with \underline{underline} and we notice that our model reaches the Top-3 performance with 10 of the 14 sub-tasks. This proves the outstanding generalize of our model.

\noindent\textbf{MMBench}\cite{MMBench} is a hand-crafted challenging benchmark, which evaluates the vision-related reasoning and perception capability with multi-choice questions. The MME Bench provides both a dev-set and test-set.
Here we report the test-set performance of our model. As shown in Table.\ref{table:mmbench_results}. Our method gets $74.4\%$ accuracy and outperforms previous methods by a large margin. Further, our model reaches the best performance in all the dimensions. This proves that our model understands the image information well and can handle diverse vision-related tasks.

\begin{table*}[t]
\vspace{-6pt}
\small
\centering
\setlength{\tabcolsep}{1mm}{
\begin{tabular}{lccccccccccccc}
\toprule
Method & Language Model &  Overall & T-Avg. & Sense.U & Inst.Id & Isnt.At & Inst.Lo & Inst.Co & Spat.R & Inst.It & Vis.R & Text.R \\
\midrule
        OpenFlamingo\cite{awadalla2023openflamingo} &  MPT-7B  & 42.7 & 39.4 & 53.2 & 45.3 & 40 & 31.2 & 39.3 & 32.6 & 36.1 & 51.4 & 25.9 \\ 
        Otter\cite{li2023otter} &  MPT-7B  & 42.9 & 40.08 & 51.3 & 43.5 & 42.3 & 34.2 & 38.4 & 30.9 & 40.2 & 55.3 & 24.7 \\ 
        MiniGPT-4\cite{zhu2023minigpt} &  Vicuna-7B  & 47.4 & 42.6 & 56.3 & 49.2 & 45.8 & 37.9 & 45.3 & 32.6 & 47.4 & 57.1 & 11.8 \\
        BLIP-2 \cite{li2022blip}&  Flan-T5-XL  & 49.7 & 45.7 & 59.1 & 53.9 & 49.2 & 42.3 & 43.2 & 36.7 & 55.7 & 45.6 & 25.9 \\ 
        IDEFICS-80b-instruct\cite{laurencon2023idefics}&  LLaMA-65B  & 53.2  & 54.4& 64 & 52.6 & 50.8 & 48.3 & 46.1 & 45.5 & 62.9 & 68 & 51.8 \\ 
        Kosmos-2\cite{peng2023kosmos} &  Kosmos 1.3B  & 54.4 & 49.4 & 63.4 & 57.1 & 58.5 & 44 & 41.4 & 37.9 & 55.7 & 60.7 & 25.9 \\ 
        InstructBLIP-Vicuna\cite{dai2023instructblip} &  Vicuna-7B  & 58.8  & 52.2 & 60.2 & 58.9 & 65.6 & 43.6 & \textbf{57.2} & 40.3 & 52.6 & 47.7 & 43.5 \\ 
        Qwen-VL-Chat\cite{bai2023qwen} &  Qwen-7B  & 65.4 & 61.9 & 73.3 & 67.3 & \textbf{69.6} & 57.7 & 52.9 & 48.2 & 59.8 & 74.6 & \textbf{53.5} \\
        LLaVA-1.5 \cite{llava1_5}& LLaMA2-7B & 66.2 & 61.8 & 74.3 & 69.9 & 67.6 & 60.3 & \textbf{57.2} & 50.8 & 63.9 & 77.3 & 35.3 \\
\rowcolor{gray!20}Ours &  InternLM-7B  & \textbf{66.9} & \textbf{65.2} & \textbf{75.0} & \textbf{71.7} & 67.6 & \textbf{60.8} & 56.2 & \textbf{55.3} & \textbf{74.4} & \textbf{77.0} & 48.5 \\ 
\bottomrule
\end{tabular}%
}
\vspace{-3mm}
\caption{
\textbf{Evaluation of Seed-Bench test set}. Here we report the results on the image-based sub tasks, including  Scene Understanding(Sense.U), Instance Identity(Inst.Id), Instance Attributes(Inst.At), Instance Localization(Inst.Lo), Instance Counting(Inst.Co), Spatial Relation(Spat.R), Instance Interaction(Inst.It), Visual Reasoning(Vis.R), Text Recognition(Text.R), and both overall accuracy(Overall) and task-level average accuracy(T-Avg.)
}
\vspace{-15pt}
\label{table:seed_results}
\end{table*}

\begin{table}[!t]
\small
\centering
\setlength{\tabcolsep}{0.7mm}{
\begin{tabular}{lcccc}
\toprule
Method & Overall &  Perception & Description & Assesment  \\
\midrule
        VisualGLM\cite{du2022glm}    & 42.4 & 53.3 & 49.8 & 24.0/23.9 \\
        Otter \cite{li2023otter}       & 46.7 & 47.2 & 60.2 & 32.3/33.3 \\
        MiniGPT-4\cite{zhu2023minigpt}    & 47.5 & 51.8 & 61.2 & 29.3/30.0 \\
        InstructBLIP\cite{dai2023instructblip} & 48.3 & 55.8 & 47.3 & 38.9/44.6 \\ 
        Shikra \cite{chen2023shikra}      & 48.3 & 55.3 & 55.7 & 33.6/34.5 \\
        LLaMA-A-V2 \cite{Gao2023LLaMAAdapterVP}  & 51.8 & 58.1 & 57.5 & 38.1/41.7 \\
        LLaVA\cite{liu2023visual}        & 54.4 & 54.7 & 62.7 & 44.4/47.4 \\ 
        LLaVA-v1.5\cite{llava1_5}   & 55.0 & 61.4 & 57.8 & 44.4/47.4 \\
        Qwen-VL-Chat\cite{bai2023qwen} & 55.6 & 61.7 & 56.0 & 47.5/50.6  \\ 
        mPLUG-Owl\cite{ye2023mplug}    & 56.4 & 58.9 & 65.7 & 43.2/45.8 \\ 
\rowcolor{Graylight}  Ours & \textbf{63.6} & \textbf{64.4} & \textbf{70.2} & \textbf{54.2/58.1} \\ 
\bottomrule
\end{tabular}%
}
\vspace{-2mm}
\caption{
\textbf{Evaluation of Q-Bench test set}. Here we report all the sub-tasks, including Perception, Description and Assessment. For Assessment, we report both SRCC and PLCC score. 
}
\vspace{-10pt}
\label{table:qbench_bench_results}
\end{table}

\noindent\textbf{Seed-Bench}\cite{li2023seedbench} is a large-scale multi-modality benchmark, which is built with the help of GPT-4 and contains nearly 19K multi-choice questions for both image and video. Here we report the image-set results in Table.\ref{table:seed_results}. It can be observed that our model gets the best overall performance and the highest performance in 6 of the 9 sub-tasks. We also notice that the sub-task data number is in-balance, so the overall metric would be biased toward some sub-tasks. 
To better evaluate the generalized capability of the LLMs along different tasks. We also report the task-level average,
and our model reaches the state-of-the-art average accuracy and outperforms the previous method with $3.3\%$. This further proves the general capability of our model.

\noindent\textbf{Q-Bench}\cite{wu2023q} is a comprehensive benchmark which focus on the low-level vision, including three realms: perception, description, and assessment. Here we report the test set results in Table.\ref{table:qbench_bench_results}. Our model shows the state-of-the-art performance on all the three tasks, surpass previous methods with a large margin. The results suggest our model has a comprehensive understanding of the image, which not only focus the high-level semantic feature of the image, but also the low-level information.

\begin{table}[!t]
\small
\centering
\setlength{\tabcolsep}{1.mm}{
\begin{tabular}{lcccccccc}
\toprule
Method & Avg & LR & AR & RR & FP-S & FP-C & CP \\
\midrule
        OpenFlamingo\cite{awadalla2023openflamingo} &  1.7 & 1.7 & 4.5 & 0 & 1.5 & 0.8 & 1.3 \\
        MiniGPT-4\cite{zhu2023minigpt} & 11.9 & 11.6 & 19.4 & 5.7 & 14.6 & 6.5 & 10.9 \\ 
        InstructBLIP\cite{dai2023instructblip} & 23.9 & 9.2 & 38.5 & 16.6 & 20.9 & 15 & 30.8 \\ 
        mPLUG-Owl\cite{ye2023mplug} &  24.9 & 6.9 & 34 & 17.5 & 33.4 & 8.5 & 30.6 \\ 
        VisualGLM\cite{du2022glm} &25.6 & 5.2 & 42 & 18 & 24.1 & 13 & 34.5 \\ 
        LLaVA\cite{liu2023visual} &  36.6 & 15 & 52.4 & 17.1 & 34.4 & 27.5 & 50.3 \\ 
        IDEFICS-80B-I\cite{laurencon2023idefics} & 38.1 & 20.8 & 49 & 27.5 & 29.1 & 36.0 & 51.2 \\
        LLaVA-1.5\cite{llava1_5} & 53.8 & 31.2 & 67.7 & 42.7 & 49.5 & 43.7 & 67.7 \\
        Qwen-VL-Chat\cite{bai2023qwen} & 56.3 & 35.3 & 63.5 & 46 & 63.6 & 43.7 & 64.7 \\ 
\rowcolor{Graylight} Ours & \textbf{72.4} & \textbf{44.5} & \textbf{79.5} & \textbf{83.4} & \textbf{71.6} & \textbf{56.3} & \textbf{82.4} \\

\bottomrule
\end{tabular}%
}
\vspace{-5pt}
\caption{
\textbf{Evaluation of MMBench-CN test set}. Here we report the results on the six L-2 abilities based on Chinese.
}
\vspace{-15pt}
\label{table:mmbench_cn_results}
\end{table}

\begin{table}[!t]
\small
\centering
\setlength{\tabcolsep}{0.7mm}{
\begin{tabular}{lcccccccc}
\toprule
Method & Avg & CP & CR & F\&C & HF & S\&B & SR & TS \\
\midrule
        OpenFlamingo\cite{awadalla2023openflamingo} &   0.7 & 1.8 & 0 & 0.8 & 0.0 & 0.0 & 2.2 & 1.5 \\
        MiniGPT-4\cite{zhu2023minigpt} &   1.7 & 7.0 & 4.0 & 0.0 & 0.0 & 1.0 & 0.0 & 0.0 \\
        LLaVA\cite{liu2023visual} &   8.3 & 10.5 & 8.1 & 7.6 & 1.7 & 8.0 & 11.1 & 10.6 \\ 
        VisualGLM\cite{du2022glm} &   9.2 & 14.0 & 11.1 & 8.4 & 0.0 & 14.0 & 4.4 & 7.6 \\
        InstructBLIP\cite{dai2023instructblip} &   12.1 & 8.8 & 9.1 & 21.0 & 0.0 & 12.0 & 6.7 & 18.2 \\ 
        mPLUG-Owl\cite{ye2023mplug} &  12.9 & 22.8 & 17.2 & 6.7 & 0.0 & 25.0 & 4.4 & 7.6 \\ 
        LLaVA-1.5\cite{llava1_5} & 16.4 & 15.8 & 19.2 & 10.9 & 3.4 & 21.0 & 37.8 & 12.1 \\
        Qwen-VL-Chat\cite{bai2023qwen} &   39.3 & 40.4 & 33.3 & 31.9 & 3.4 & \textbf{67.0} & 51.1 & 42.4 \\ 
\rowcolor{Graylight} Ours &   \textbf{47.6} & \textbf{50.9} & \textbf{53.5} & \textbf{42.0} & \textbf{10.3} & 55.0 & \textbf{73.3} & \textbf{50.0} \\ 
\bottomrule
\end{tabular}%
}
\vspace{-3mm}
\caption{
\textbf{Evaluation of CCBench test set}. We report all the sub-tasks, including Calligraphy Painting(CP), Cultural Relic(CR), Food \& Clothes(F\&C), Historical Figures(H\&F), Scenery \& Building(S\&B), Sketch Reasoning(SR), Traditional Show(TS), 
}
\vspace{-15pt}
\label{table:chinese_bench_results}
\end{table}

\subsection{Chinese-Based Benchmark results.}
As we introduced in Sce.\ref{sec:intro}, our model is pretrained with rich multilingual knowledge. To prove the effectiveness of the pretraining, here we further evaluate its performance with two Chinese-based benchmarks.

\noindent\textbf{MMBench-CN}\cite{MMBench} is the Chinese translated benchmark of the original MMbench, which shows the vision-related Chinese understanding and reasoning capability. 
Here we report the test-set performance in Table.\ref{table:mmbench_cn_results}. It can be observed that our method outperforms previous methods by a large margin. When comparing with the English version performance in Table.\ref{table:mmbench_results}. Qwen and VisualGLM have $4.9\%$ and  $7.9\%$ performance degrading, while the performance gap of our model between different languages is only $2.0\%$. This proves the strong multi-lingo capability of our model.

\noindent\textbf{Chinese-Bench}\cite{MMBench} is a Chinese knowledge-related benchmark, that challenges the model with Chinese traditional cultural questions, including art, food, clothes, landmarks, \etc.  
The performance is given in Table.\ref{table:chinese_bench_results}. It can be observed that the benchmark is quite challenging, most LLaMA-based model fails to answer these questions, due to the lack of corresponding knowledge. Compared with LLaMA-based methods, the Qwen-based model Qwen-VL-Chat shows a much better performance of $39.3\%$. While it is still worse than InternLM-based model, which reaches a new state-of-the-art performance of $47.6\%$. This proves the rich Chinese knowledge of IntenrLM and the great alignment between the vision and language knowledge by our large-scale pre-training.

\begin{table}[!t]
\small
\centering
\setlength{\tabcolsep}{0.8mm}{
\begin{tabular}{l|cccc|cccc}
\toprule
       & \multicolumn{4}{c|}{Human Evaluation} & \multicolumn{4}{c}{GPT4-Vision Evaluation}  \\
Method & Avg. & Pref. & Text & Image & Avg. & Pref. & Text & Image  \\
\midrule
Qwen-14B\cite{qwen7b}     & 0.62 & 0.52 & 0.80 & 0.54 & 
0.82 & 0.78  & 0.91 & 0.72 \\ 
Intern-20B\cite{2023internlm} & 0.64 & 0.54 & 0.82 & 0.56 &
0.81 & 0.82  & 0.89 & 0.70 \\ 
GPT3.5  \cite{openai2020chatgpt}     & 0.71 & 0.64 & 0.88 & 0.59 &
0.84 & 0.84  & 0.91 & 0.75  \\ 
\rowcolor{Graylight}  Ours & \underline{0.74} & \underline{0.65}   & \underline{0.90} & \textbf{0.67} & 
\underline{0.87} & \textbf{0.90} & \underline{0.93} & \textbf{0.79} \\
GPT4-V \cite{openai2023gpt4}        & \textbf{0.77} & \textbf{0.73} & \textbf{0.95} & \underline{0.64} & 
\textbf{0.89} & \underline{0.88} & \textbf{0.96} & \underline{0.78} \\  
\bottomrule
\end{tabular}%
}

\vspace{-3mm}
\caption{
Quantitative results by human and GPT-4Vision. \textbf{Text} means the writing-related scores and  \textbf{Image} means the image-selection related scores. 
The evaluation by human and GPT-4Vision shows consist conclusion that GPT-4 performs best and our model is on par with the GPT3.5, better than other LLMs.
}
\vspace{-3mm}
\label{table:user_study}
\end{table}

\begin{table}[!t]
\small
\centering
\setlength{\tabcolsep}{3.5mm}{
\begin{tabular}{lcccc}
\toprule
Method & Human & Top-1 & GPT4-Vision & Our \\
\midrule
Score & 100.0 & 40.2 & 51.7 & 80.4 \\
\bottomrule
\end{tabular}%
}
\vspace{-3mm}
\caption{
Effectiveness of Image Selection, \textbf{Human} is the ground truth selected by human.  
}
\label{table:image_selection}
\vspace{-5mm}
\end{table}

\subsection{Interleaved Image-Text Composition}

\noindent\textbf{Quantitative results.}
Due to the lack of existing metrics to evaluate the image-text interleaved article quality, we designed a detailed user study metric\footnote{Please refer to the supplementary materials for more details.}, including eight dimensions: four text-related dimensions (instruction following, writing quality, logic, factualness), three image-related dimensions (image-text consistency, image informative, image consistency), and one subjective preference. As our InternLM-XComposer is the first interleaved image-text composition model, we compare the composition quality with recent leading LLMs and VLMs, including GPT4-V, GPT3.5, InternLM-20B, and Qwen-14B. For a fair comparison, we apply a similar composition pipeline as mentioned in Sec.\ref{sec:compos_data} in these solutions. Notably, the image selection phase is not supported in language-only models (GPT3.5, InternLM-20B, and Qwen-14B), in which retrieved images with top-1 similarity are adopted as the final illustrations. In practice, we generated articles with 20 articles. For robustness and reliability of final results, we adopt both human evaluation and GPT4-V scoring. Specifically, we invite 10 human experts for assessment. In the meantime, GPT4-V also rates scores for these articles on the proposed eight dimensions.

As presented in Table.\ref{table:user_study}, GPT-4V shows the best performance in all the text-related dimensions and the highest overall score. Benefiting from the interleaved image-text composition training, our InternLM-XComposer performs the best on image-related dimensions and the second best on the text-related dimensions and average score.

\noindent\textbf{Effectiveness of Image Selection}
The last step of our interleaved Image-Text composition pipeline is the Image retrieval and selection. The InternLM-XComposer could select a proper image from the retrieved candidates based on the context. Here we study the effectiveness of the image selection based on human preference. Specifically, we provide the text and candidate images to human experts, GPT4-Vision and Intern-XComposer, and use the human-selected images as the ground truth. We also use the image with the highest retrieval similarity as the `Top-1' baseline.
As shown in Table.\ref{table:image_selection}, InternLM-XComposer gets a higher score than GPT4- V, which means the image selected by our model is more consistent with the human preference.

\noindent\textbf{Qualitative results}. We direct readers to the supplementary materials for detailed qualitative results of the interleaved image-text compositions and multimodal conversations generated by the InternLM-XComposer.

\subsection{Ablation Study}
\noindent\textbf{Learnable Components}.
Here we study the influence of different learnable components in the multi-task supervised fine-tuning stage. As shown in Table.\ref{table:component_ablation}, the Perceive Sampler is critical for the multi-task learning, which shows significant influence on most benchmarks. For the LoRA in LLM, the Attention and FFN part are also important to realize a superb performance.

\begin{table}[!t]
\small
\centering
\setlength{\tabcolsep}{1.7mm}{
\begin{tabular}{lccc}
\toprule
Method & MME & MMBench & Seed Bench \\
\midrule
  
\rowcolor{gray!20}Ours & 1919.5 & 74.4 & 66.9  \\ 
\midrule
Freeze Perceive Sampler &  1819.1 & 71.3 & 63.2 \\
Freeze Attention LoRA & 1776.5 & 72.0 & 65.9\\
Freeze FFN LoRA & 1828.7 & 72.4 & 66.4\\
\bottomrule
\end{tabular}%
}
\vspace{-3mm}
\caption{
Ablation on Learnable Components.  
}
\vspace{-3mm}
\label{table:component_ablation}
\end{table}

\begin{table}[!t]
\small
\centering
\setlength{\tabcolsep}{1mm}{
\begin{tabular}{lcccc}
\toprule
LLM Backbone & MME & MMBench & MMBench-CN & CCBench \\
\midrule
    
LLaMA-2 \cite{touvron2023llama2} & 1895.9 & 72.6 & 66.7 & 44.6 \\
\rowcolor{gray!20}
InternLM\cite{2023internlm} & 1919.5 & 74.4 & 72.4 & 47.6 \\ 
\bottomrule
\end{tabular}%
}
\vspace{-3mm}
\caption{
Ablation on LLM Backbone.
}
\vspace{-5mm}
\label{table:backbone_ablation}
\end{table}

\noindent\textbf{LLM selection}.
Then we study the influence of different LLMs, here we consider the InternLM-7B and LLaMA2-7B, with the same pre-training and supervised fine-tuning strategy. Here we report the results on MME, MMBench, MMBench-CN and CCBench. With the results in Table.\ref{table:backbone_ablation}, we find the LLaMA-2 performs similar with InternLM-7B in the English-based benchmarks, while the performances gap is enlarged in the Chinese-based benchmarks.

\section{Conclusion}
In this paper, we present InternLM-XComposer, a vision-language large model with superb multi-modality understanding and composition capability. Benefiting from the rich multi-lingual and multi-modality knowledge from carefully designed pretraining, our model could generate coherent interleaved image-text composition, and shows state-of-the-art performance across various vision-language LLM benchmarks. We hope XComposer could provide new insight for the advanced vision-language interaction.

{\small
\bibliographystyle{ieee_fullname}
\bibliography{egbib}
}

\renewcommand{\thetable}{A\arabic{table}}
\renewcommand{\thefigure}{A\arabic{figure}}

\newpage
\clearpage
\appendix
\newpage
\label{sec:supp}
\section{Experiment Details}
\subsection{Pre-training }

The multi-modal pre-training process employs 1.1 billion images alongside 77.7 billion text tokens, including both public datasets and in-house concept data (Sec.~\ref{sec:conceptdata}) collected from public websites, possessing over 11 million semantic concepts. The paired and interleaved text-image data includes 50.6 billion English text tokens and 17.1 billion Chinese text tokens. Furthermore, approximately 10 billion text tokens (5 billion English text and 5 billion Chinese text), sampled from the InternLM pre-training dataset~\cite{He2023WanJuanAC}, are incorporated to maintain the model's linguistic proficiencies. Prior to the training process, all pre-training data underwent a thorough cleaning procedure to ensure its quality and reliability.

The optimization algorithm employed is AdamW, with hyperparameter settings as follows: $\beta_1$=0.9, $\beta_2$=0.95, $eps$ =1e-8. The maximum learning rates for the perceive sampler and the large language model are configured at 2e-4 and 4e-5, respectively, following a cosine learning rate schedule. The minimum learning rate is set at 1e-5. Additionally, a linear warm-up is applied over the initial 200 steps. The training procedure employs a batch size of approximately 15.7 million tokens and spans 8,000 iterations. Utilizing such a large batch size in conjunction with a limited number of iterations contributes to stable training dynamics while also aiding in the preservation of LLM's inherent capabilities. The overall multi-modal pre-training process requires 128 Nvidia A100 GPUs for around 80 hours.

\subsection{Supervised Fine-tuning }
We adopt the Low-Rank Adaption (LoRA)~\cite{hu2022lora} for the supervised fine-tuning process, which is composed of a \emph{Multi-task Training} phase and a \emph{Instruction Tuning} phase. The perceive sampler is concurrently trained, albeit with a distinct learning rate. Specifically, LoRA is applied to both the attention layer and the feed-forward network (FFN). We find that a high LoRA rank is conducive to imbuing the model with new capabilities; consequently, we set the LoRA rank and alpha parameter both to 256. The model is trained using a global batch size of 256 over 18,000 iterations in \emph{Multi-task Training} phase. The learning rates are set to $5e^{-5}$ for the LoRA layer and $2e^{-5}$ for the perceive sampler.

For the \emph{Instruction Tuning} phase, we maintain a batch size of 256 and execute the tuning over 1000 iterations with a small learning rate $1e^{-5}$.

\section{Concept Data}
\label{sec:conceptdata}
\begin{figure}[t]
	\centering
	\includegraphics[width=0.95\linewidth]{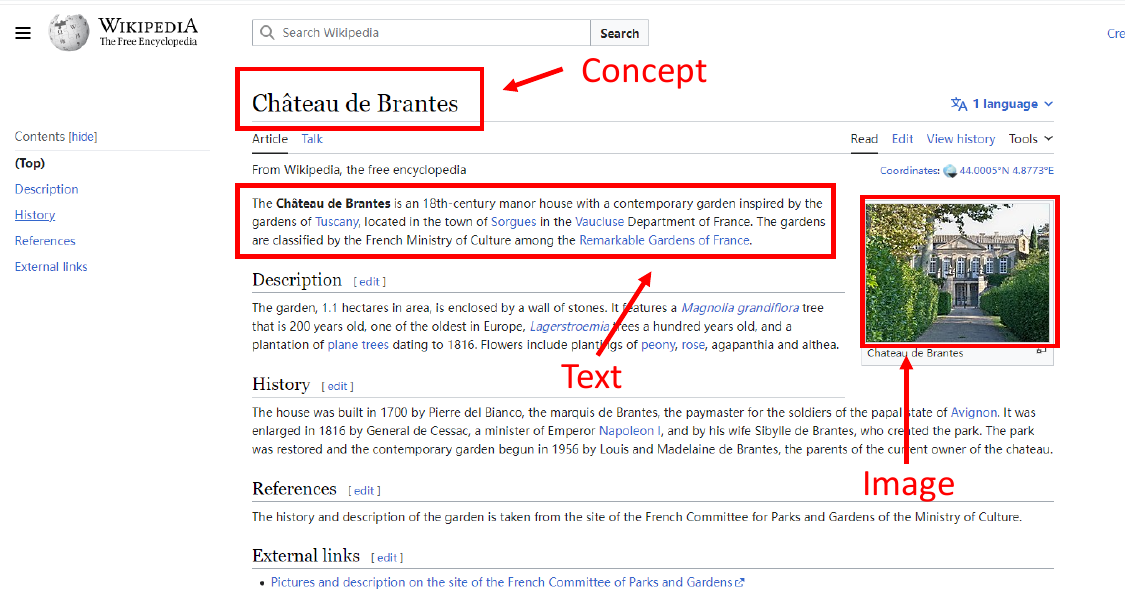}
 \vspace{-5pt}
    \caption{Extract vision-language concepts from Wikipedia.}
	\label{fig:Wikipedia}
 \vspace{-15pt}
\end{figure}

\begin{figure}[t]
	\centering
	\includegraphics[width=0.92\linewidth]{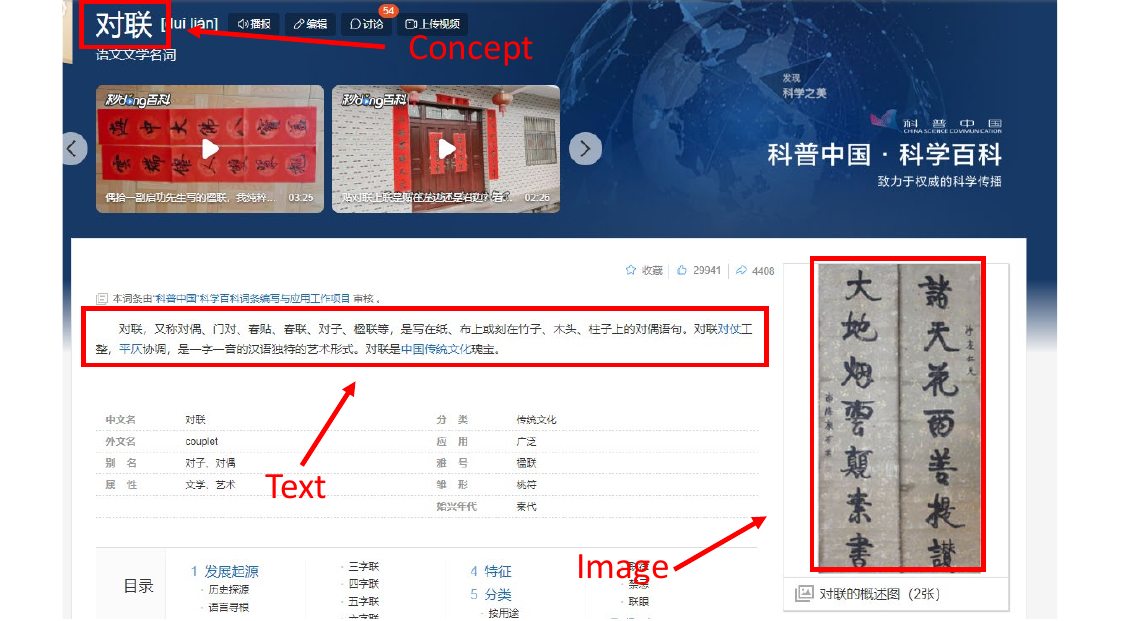}
\vspace{0pt}
    \caption{Extract vision-language concepts from Baidu Baike.}
	\label{fig:baidubaike}
  \vspace{-10pt}
\end{figure}

\begin{figure*}[t]
	\centering
	\includegraphics[width=0.95\linewidth]{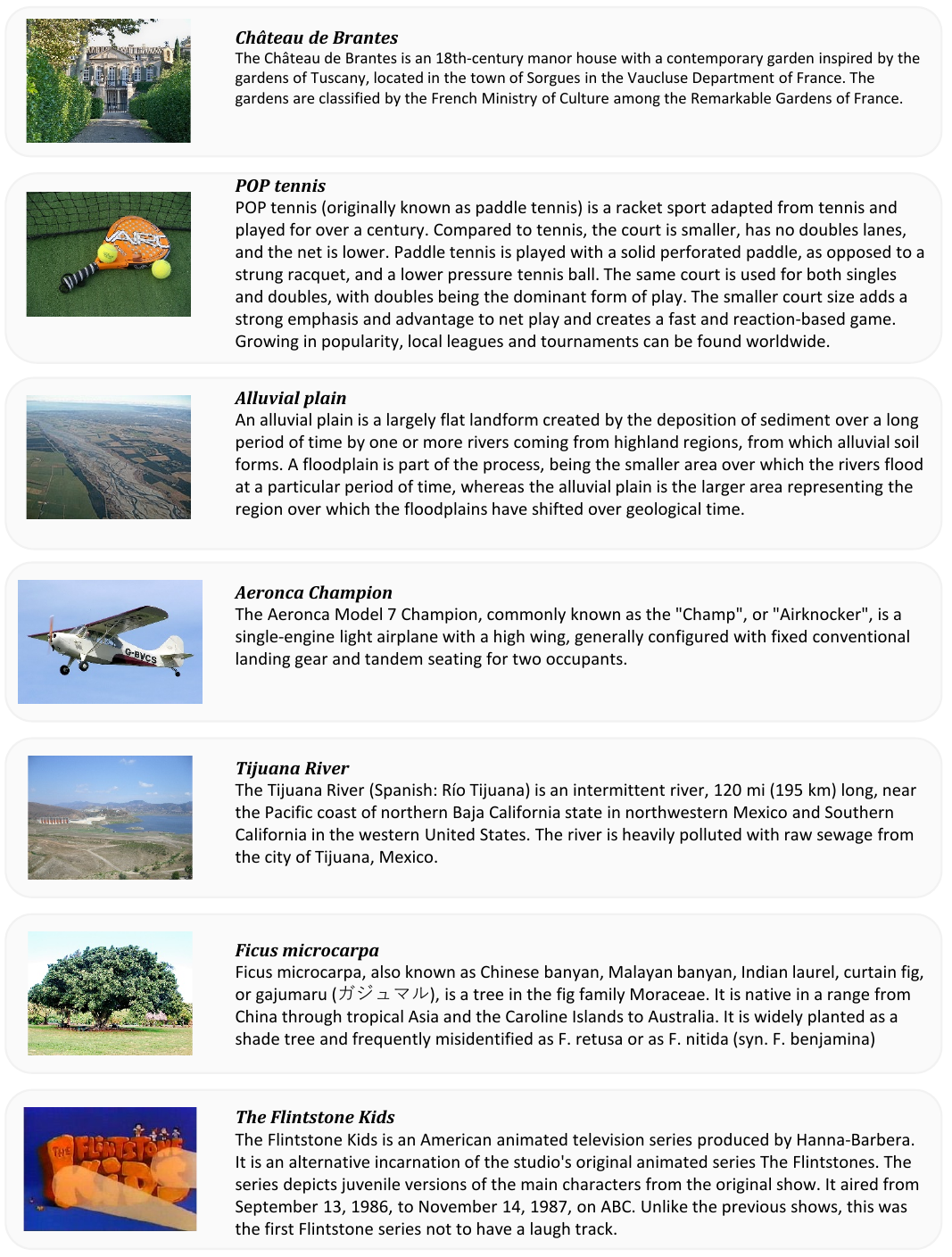}
    \caption{Examples from our vision-language concept dataset.}
	\label{fig:concepts1}
\end{figure*}

\begin{figure*}[t]
	\centering
	\includegraphics[width=0.95\linewidth]{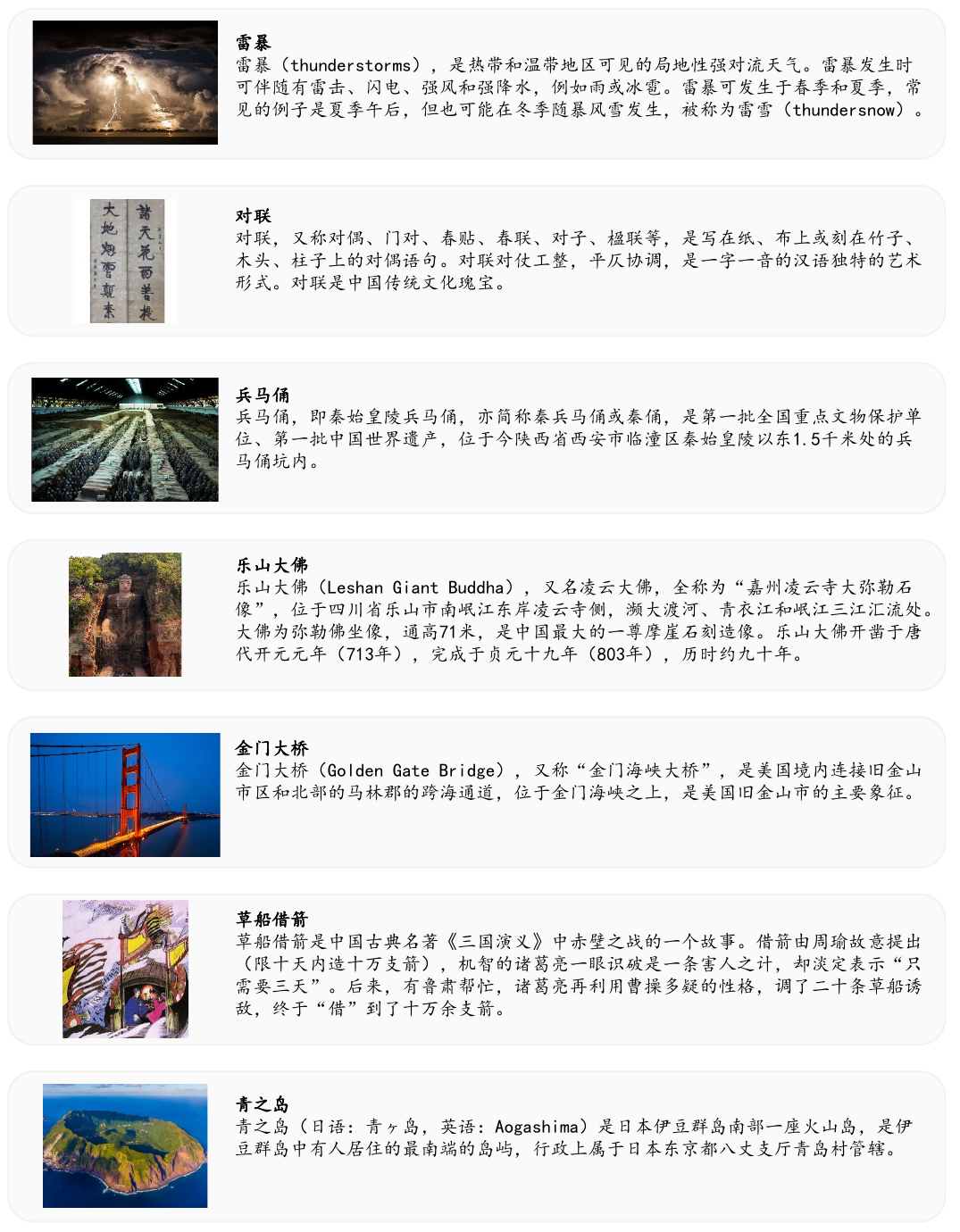}
    \caption{Examples from our vision-language concept dataset.}
	\label{fig:concepts2}
\end{figure*}

As illustrated by Figure~\ref{fig:Wikipedia} and Figure~\ref{fig:baidubaike}, our concept dataset is curated from Wikipedia~\footnote{https://www.wikipedia.org/} and Baidu Baike~\footnote{https://baike.baidu.com/}, comprising 2 million English vision-language concepts and 9 million Chinese vision-language concepts. Each concept consists of an image paired with corresponding descriptions. Examples are illustrated in Figure~\ref{fig:concepts1} and Figure~\ref{fig:concepts2}. In comparison to the WIT~\cite{srinivasan2021wit} dataset, which comprises approximately 2 million multimodal English and Chinese concepts after removing expired links and noisy samples, our multimodal concept data significantly surpasses it in terms of data volume and diversity. 

\noindent\textbf{Experimental results.}
Here, we study the effectiveness of the concept dataset on our framework with three settings: 1) remove our In-house Concept data.  2) replace our In-house Concept data with the WIT~\cite{srinivasan2021wit} dataset. 3) the default pre-training setting.
As results shown in Table~\ref{table:concept_data_ablation}, compared to the baseline, the WIT improves the model performance on both English and Chinese benchmarks. Further, when using our larger and more diverse In-house Concept dataset, the model surpasses the WIT baseline by a large margin on both benchmarks, especially on the Chinese knowledge-based benchmark CCBench.

\section{Interleaved Image-Text Composition}
\subsection{Format Details}
\noindent{\textbf{Text Generation.}}
As the interleaved image-text dataset is collected from public websites, the acquired dataset contains noise, particularly in the form of marketing and advertising content. To address this, 
We utilize GPT-4 to assess whether an individual sentence contains noise and identify the type of noise, which includes advertisement, reference, and recommendations (\eg, \textit{external articles/answers}). 
Any sentences identified as noisy are removed, and articles containing more than 30\% noisy sentences are filtered out directly.
After the cleaning, we formulate the training data in the following manner:
\begin{align*}
& \texttt{<|User|>}: \textit{Write an illustrated article based on the} \\
& \quad\quad\quad\quad\quad\quad \textit{given instruction: \{Instruction\}} \enspace \texttt{<eou>}  \\
& \texttt{<|Bot|>}: [para_1]\ldots[para_N] \enspace \texttt{<eob>}
\end{align*}
Here, $\{Instruction\}$ serves as a placeholder for the article-crafted instruction, such as a simple title, or a draft outline, \etc. The $[para_1]$ and $[para_N]$ denote the first and last paragraphs, respectively.

\noindent{\textbf{Image Spotting and Captioning.}}
In practice, we formulate the spotting and captioning task with the following format:
\begin{align*}
& \texttt{<|User|>}: [seg_1][para_1]\ldots[seg_N][para_N] \enspace \textit{Based}\\
& \quad\quad\quad\quad\quad\quad \textit{on the above text, select the appropriate} \\
& \quad\quad\quad\quad\quad\quad \textit{locations for inserting images and give} \\
& \quad\quad\quad\quad\quad\quad \textit{the image captions} \enspace \texttt{<eou>}  \\
& \texttt{<|Bot|>}: \textit{I think images can be added under the \{$x_1\},$}\\
& \quad\quad\quad\quad\quad \textit{$\ldots, \{x_k\}$ paragraphs. The image captions are} \\
& \quad\quad\quad\quad\quad \textit{as follows: \{$x_1$\} paragraph: \{$cap_1$\}, $\ldots$}, \\
& \quad\quad\quad\quad\quad \textit{ \{$x_k$\} paragraph: \{$cap_k$\}} \enspace \texttt{<eob>}
\end{align*}
Here, \( [seg_1] \) serves as an index token to pinpoint the specific paragraph index. The placeholders \( \{x_1\} \) and \( \{x_k\} \) represent the positions for the first and last image locations, respectively. Correspondingly, \( \{cap_1\} \) and \( \{cap_k\} \) act as the generated captions associated with those image locations.

\begin{table}[t]
\small
\centering
\setlength{\tabcolsep}{3mm}{
\begin{tabular}{ccc}
\toprule
 Concept-specific Data & MMBench & CCBench \\
\midrule
-  & 72.9 & 44.0 \\
WIT (2M) &  73.5 & 45.1 \\
\rowcolor{gray!20}
In-house Concept data (11M) & 74.4 & 47.6 \\  
\bottomrule
\end{tabular}%
}
\vspace{-5pt}
\caption{
Ablation of the concept-specific data used in the pre-training phase.
}
\vspace{-30pt}
\label{table:concept_data_ablation}
\end{table}

\noindent{\textbf{Image Retrieval and Selection.}}
The training data is structured in the following manner:
\begin{align*}
& \texttt{<|User|>}: [para_1]\ldots[para_i][img_i][para_{i+1}]\ldots[para_j] \\
& \quad\quad\quad\quad\quad\quad \textit{Based on the given context and candidate} \\
& \quad\quad\quad\quad\quad\quad \textit{images, select the appropriate image.} \\
& \quad\quad\quad\quad\quad\quad \textit{Candidate images include:} [img_j^1]\ldots[img_j^m]\enspace \\
& \quad\quad\quad\quad\quad\quad \texttt{<eou>}  \\
& \texttt{<|Bot|>}: \textit{The \{selected index\} image.} \enspace \texttt{<eob>}
\end{align*}
In this configuration, \( [img_i] \) denotes the image associated with the \( i^{th} \) paragraph.
The terms \( [img_j^1], \ldots, [img_j^m] \) represent the candidate images retrieved by CLIP~\cite{radford2021learning} based on \( \{cap_1\} , \ldots, \{cap_k\} \) from an image pool. Meanwhile, \( \{\textit{selected index}\} \) acts as a placeholder indicating the index of the final selected image.

\subsection{Image-Text Composition Evaluation Details}
\noindent{\textbf{Grading Criteria.}}
Due to the lack of existing metrics to evaluate the interleaved image-text article quality, we designed a detailed user study metric, including eight dimensions: 
four text-related dimensions (instruction following, writing quality, logic, factualness), three image-related dimensions (image-text consistency, image informative, image consistency), and one subjective preference. Each dimension contains three levels of score: Excellent (5 points), Medium (3 points), and Poor (1 point). 

With a given topic, the human expert and GPT4-V are asked to grade the articles generated by different models (in a double-blind manner) in each of the eight dimensions. Specifically, for each time of scoring one article, the points of four text-related dimensions are summarized and divided by the full score of text (20 points) into a text-related score, and the three image-related dimensions are summarized and divided by the full score of image (15 points) into an image-related score. The subjective preference is divided by five into a subjective preference score. The average score, which is the average number of the text-related score, the image-related score, and the subjective preference score, is taken as the overall result for a human expert or the GPT4-V. The scores of 10 human experts are averaged for the final human evaluation result. The GPT4-V assessments are performed three times to calculate the average GPT4-V score.

\noindent{\textbf{Details of Eight Evaluation Dimensions.}} The details of each of the eight dimensions are as follows:

\noindent\underline{\emph{1) Instruction following}} evaluate the content based on the degree of task completion, and evaluate whether the generated article conforms to the given topic and meets the style required by the topic (such as meeting the specified title, complying with the given abstract, keywords, outline, etc.). Scoring Levels: 

1 point: Severely non-compliant with instructions; 

3 points: Basically compliant with instructions; 

5 points: Highly compliant with instructions.

\noindent\underline{\emph{2) Writing quality}} evaluate the language quality based on the content of the article (including but not limited to vocabulary, grammar, word choice and sentence making, etc.). It requires accurate word expression, rich language, and correct grammar. Scoring Levels: 

1 point: Poor; 

3 points: Medium; 

5 points: Excellent.

\noindent\underline{\emph{3) Logic}} evaluate the article based on the logical rationality of the content (the content of the article should try to follow the common rules in daily life, have clear cause and effect relationships, and have no obvious logical fallacies, etc.). Scoring Levels: 

1 point: Severe logical errors (3 or more instances); 

3 points:  Minor logical errors (1-3 instances); 

5 points: Clear logic and explicit cause-and-effect relationship.

\noindent\underline{\emph{4) Factualness}} evaluate the article based on the accuracy of the content (the content of the article should be as consistent as possible with common sense of life and scientific knowledge, without illusions or fabricated facts, and the use of relevant documents and materials should be reasonable and appropriate). Scoring Levels: 

1 point: Severe factual errors or delusions (3 or more instances); 

3 points: Minor factual errors or delusions (1-3 instances); 

5 points: Factually correct, with no obvious delusions or fabrications.

\begin{table}[t]
\footnotesize
\centering
\begin{tabular}{l}
    \toprule
    \textbf{GPT4-Vision Evaluation Prompt}
    \\
    \midrule
    There is an interleaved text-image article that should follow the instruction: \\
    \{\textit{Instruction}\} \\
    The article's content: \\
    \{\textit{Article content}\} \\
    Based on the given instruction and article, please evaluate this article by \\
    answering the following questions: \\
    Please evaluate whether the article meets the requirements and \\
    choose the score from the given scoring level in each question: \\
    Q1: $\{Q1\}$ \\
    $\ldots$ \\
    Q8: $\{Q8\}$ \\
    Please give a rating for each question one by one and explain the reason. \\
    The output format should be: \\
    "Final Score: \\
    Q1: (1 or 3 or 5) points, Reason: xxx \\
    $\dots$ \\
    Q8: (1 or 3 or 5) points, Reason: xxx "\\
    \bottomrule
\end{tabular}
\vspace{-5pt}
\caption{The prompt used for GPT4-V evaluation.  $\{\textit{Instruction}\}$, $\{\textit{Article content}\}$, and $\{Q1\}\ldots\{Q8\}$ will be replaced by the instruction, the interleaved image-text article, and explanations and scoring levels of eight dimensions, respectively.}
\label{tab:gpt4v_prompt}
\vspace{-10pt}
\end{table}

\noindent\underline{\emph{5) Image-text consistency}} evaluate the relevance of the images to the topic of the article. Scoring Levels: 

1 point: The illustration is unrelated to the article's theme; 

3 points: The illustration is related to the article's theme, but not closely; 

5 points: The illustration is closely related to the article's theme.

\noindent\underline{\emph{6) Image informative}} evaluate the relevance of the main content of the accompanying images and related paragraphs, and the informativeness of the images. Scoring Levels: 

1 point: The illustration is completely uninformative; 

3 points: The illustration is related to the main content of the relevant paragraph; 

5 points: In addition to the above, the illustration also conveys extra supplementary information not present in the article.

\noindent\underline{\emph{7) Image consistency}} evaluate the subject consistency between the accompanying pictures (for instance, when writing about my cat, the accompanying image should ideally be of the same cat.). Scoring Levels: 

1 point: The subjects in some illustrations differ greatly; 

3 points: The subjects in some illustrations appear similar, but upon closer examination, differences exist; 

5 points: Completely satisfies image consistency.

\noindent\underline{\emph{8) Subjective preference}} is an overall subjective score about the quality of the generated article. Scoring Levels: 

1 point: Poor; 

3 points: Medium; 

5 points: Excellent.

\noindent{\textbf{GPT4-V Evaluation.}} Besides human experts, we leverage GPT4-V api, which can take interleaved text-image articles as inputs for evaluation. The input prompt format is shown in Table~\ref{tab:gpt4v_prompt}. $\{\textit{Instruction}\}$, $\{\textit{Article content}\}$ will be replaced by the requirements of the article and the composed interleaved image-text article during evaluation, respectively. $\{Q1\}\ldots\{Q8\}$ indicates the explanations and scoring levels of eight dimensions as in \textbf{Details of Eight Evaluation Dimensions}, \eg, "Instruction following evaluate the content $\ldots$ Scoring Levels: $\ldots$".
We manually check the output of the GPT4-V api to collect the final score. 

\begin{table}[!t]
\small
\centering
\setlength{\tabcolsep}{1.5mm}{
\begin{tabular}{lcccccc}
\toprule
Method & Overall &  VR  & VP & VKA & VC & OH  \\
\midrule
        MiniGPT-4~\cite{zhu2023minigpt}    & 192.6  &  37.6 & 37.8 & 17.6 & 49.0 & 50.7 \\
        LLaVA~\cite{liu2023visual}        & 197.0  &  41.6 & 38.3 & 18.7 & 49.4 & 49.0 \\ 
        VisualGLM~\cite{du2022glm}    & 211.9  &  37.3 & 36.3 & 46.9 & 37.6 & 54.0 \\
        Otter~\cite{li2023otter}        & 216.4  &  41.6 & 37.0 & 15.1 & 52.4 & 74.0 \\
        LLaMA-A-V2~\cite{Gao2023LLaMAAdapterVP}   & 229.2  &  43.5 & 46.8 & 22.3 & 56.0 & 60.7 \\
        Lynx~\cite{Zeng2023WhatMI}         & 279.2  &  52.2 & \underline{65.8} & 17.6 & 57.4 & {86.3} \\
        BLIP2~\cite{Li2023BLIP2BL}        & 284.7  &  44.9 & 49.0 & \underline{64.1} & 44.0 & 82.7 \\
        InstructBLIP~\cite{dai2023instructblip} & 300.6  &  46.7 & 48.0 & 61.7 & \underline{59.2} & 85.0 \\
        LLaVA-1.5~\cite{llava1_5}    & 307.2  & 55.6 & 49.0 & 57.0 & 57.2 & \underline{88.3}\\
        Qwen-VL-Chat~\cite{bai2023qwen} & 316.8  &  \underline{62.4} & \underline{54.5} & 55.1 & 54.8 & \underline{90.0}  \\
        Bard~\tablefootnote{https://bard.google.com/. \\ Refer to Tiny LVLM at https://github.com/OpenGVLab/Multi-Modality-Arena/tree/main/tiny\_lvlm\_evaluation} & 319.6  &  \underline{64.2} & \underline{57.0} & \underline{68.1} & \underline{59.6} & 70.7  \\  
\rowcolor{gray!20} Ours & \textbf{322.5} & \underline{55.8} & 53.8 & \underline{64.1} & \underline{61.8} & \underline{87.0}  \\ 
\bottomrule
\end{tabular}%
}
\caption{
\textbf{Evaluation of Tiny LVLM test set}. Here we report all the sub-tasks, including visual reasoning (VR), visual perception (VP), visual knowledge acquisition (VKA), visual commonsense (VC), and object hallucination (OH).
}
\label{table:lvlm_bench_results}
\vspace{-10pt}
\end{table}

\section{More Benchmark results.}

\noindent\textbf{Tiny LVLM}~\cite{xu2023lvlmehub} is an ability-level benchmark, which evaluates the MLLM performance from five different abilities. We report the results in Table~\ref{table:lvlm_bench_results}. Our InternLM-Xcomposer-VL gets the best overall results and the Top-3 performance in most abilities, even surpassing a commercial MLLM, \ie, Google Bard.

\section{More Qualitative Results.} \label{app:vis}
\begin{figure*}[t!]
	\centering
	\includegraphics[width=0.95\linewidth]{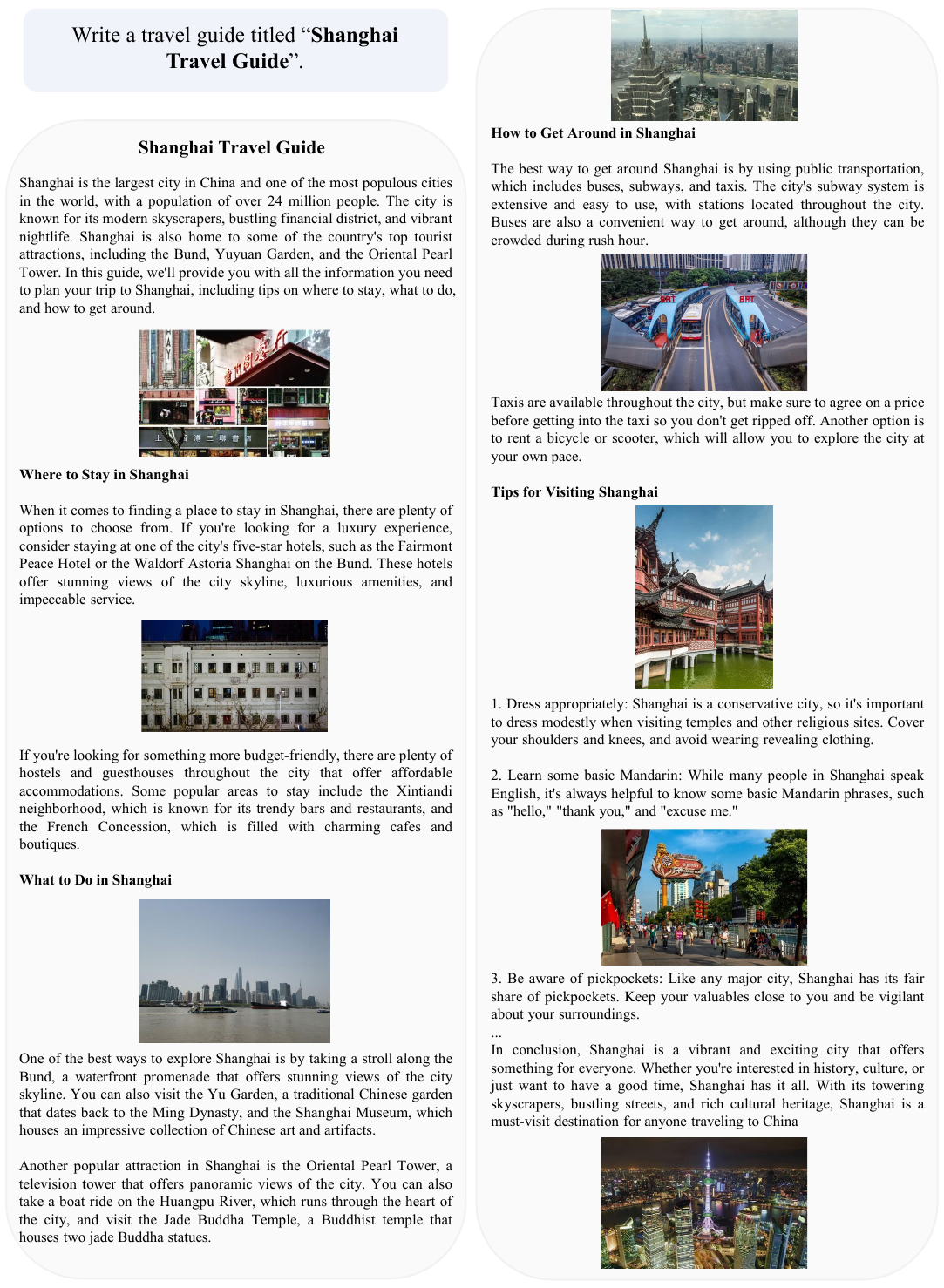}
    \vspace{-10pt}
    \caption{Interleaved composition generated by InternLM-XComposer.}
	\label{fig:article1}
\end{figure*}

\begin{figure*}[t!]
	\centering
	\includegraphics[width=0.95\linewidth]{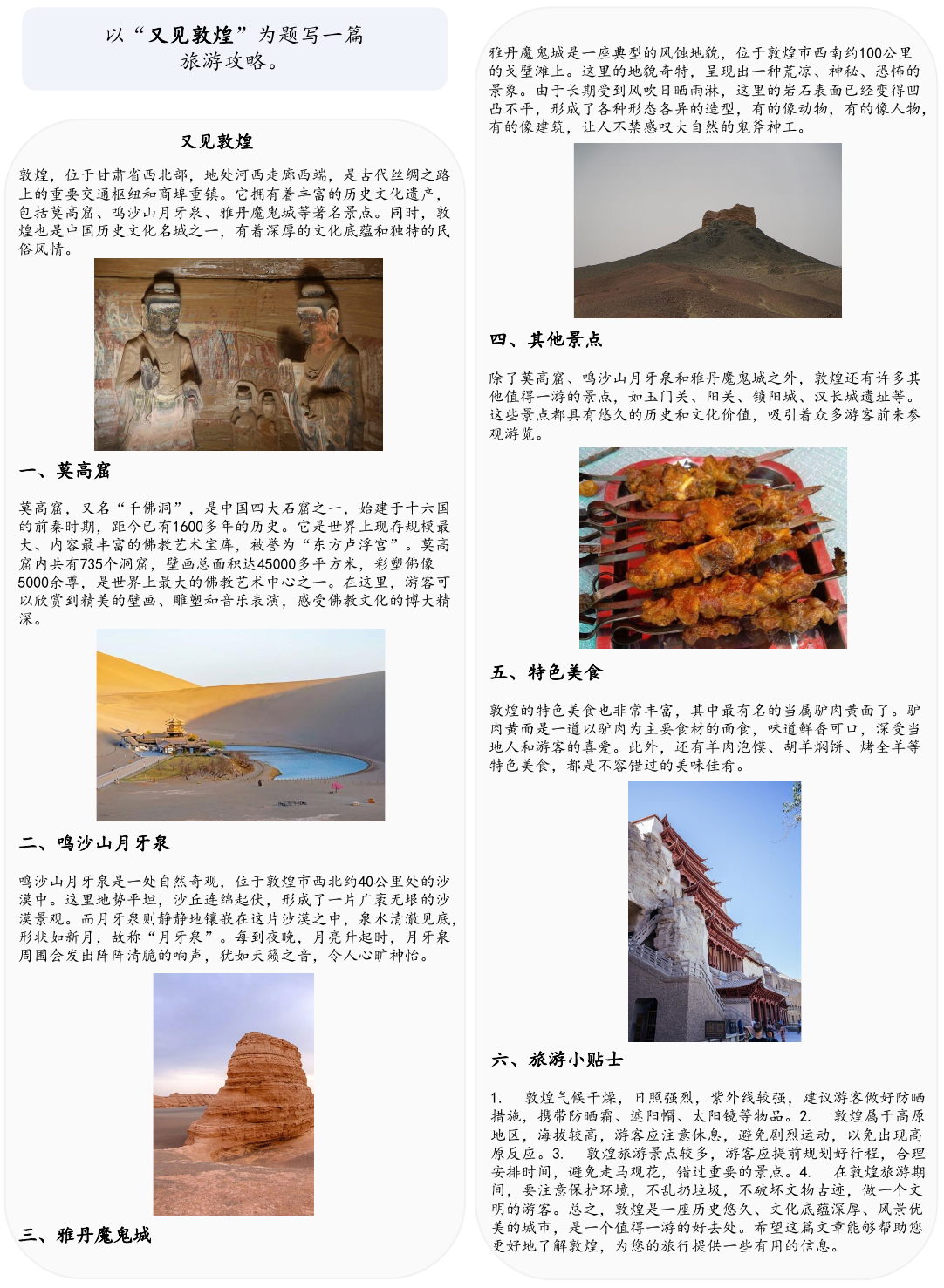}
    \vspace{-10pt}
    \caption{Interleaved composition generated by InternLM-XComposer.}
	\label{fig:article2}
\end{figure*}

\begin{figure*}[t!]
	\centering
	\includegraphics[width=0.95\linewidth]{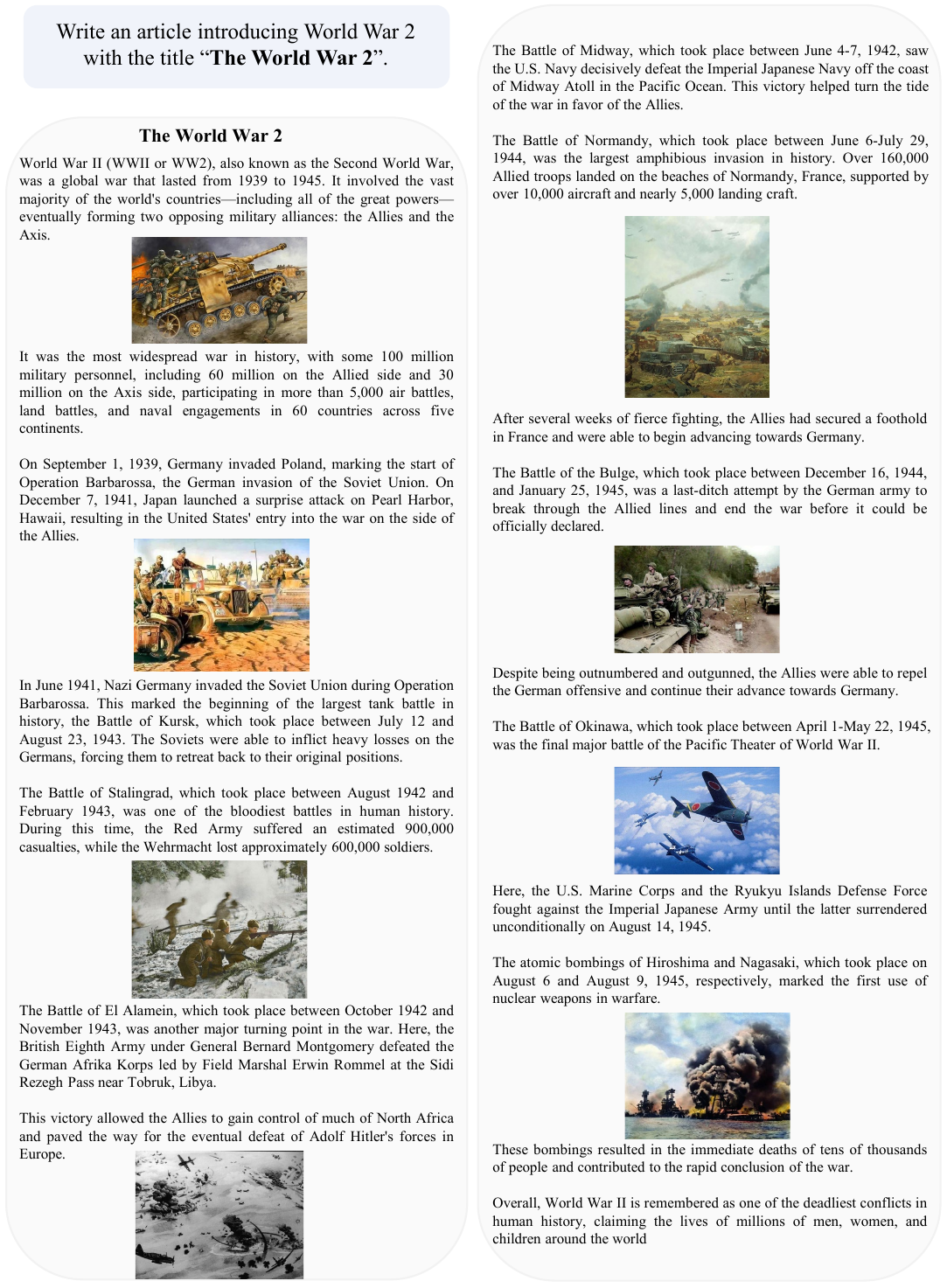}
    \vspace{-10pt}
    \caption{Interleaved composition generated by InternLM-XComposer.}
	\label{fig:article3}
\end{figure*}

\begin{figure*}[t!]
	\centering
	\includegraphics[width=0.95\linewidth]{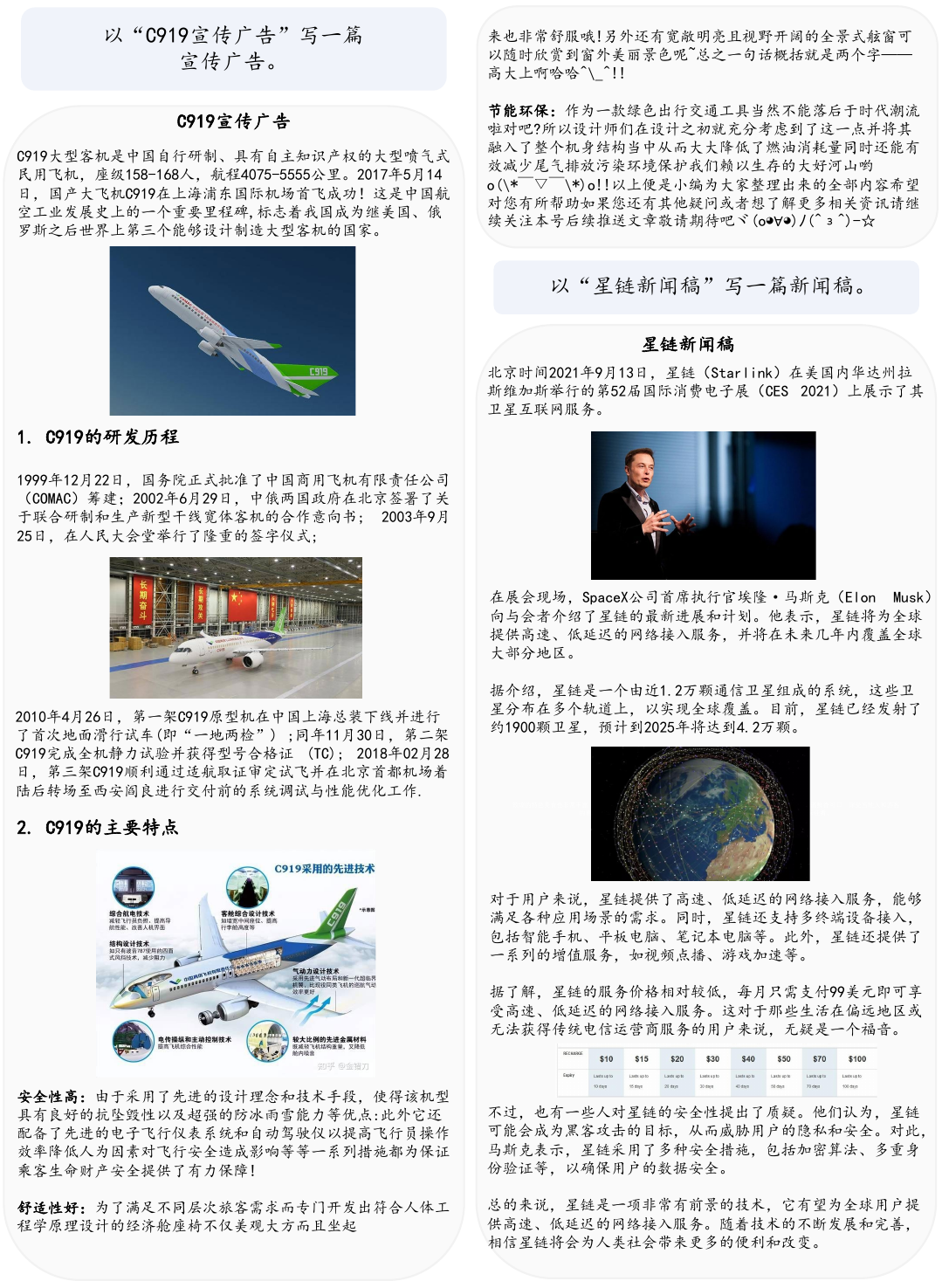}
    \vspace{-10pt}
    \caption{Interleaved composition generated by InternLM-XComposer.}
	\label{fig:article4}
\end{figure*}

\begin{figure*}[t]
	\centering
	\includegraphics[width=0.95\linewidth]{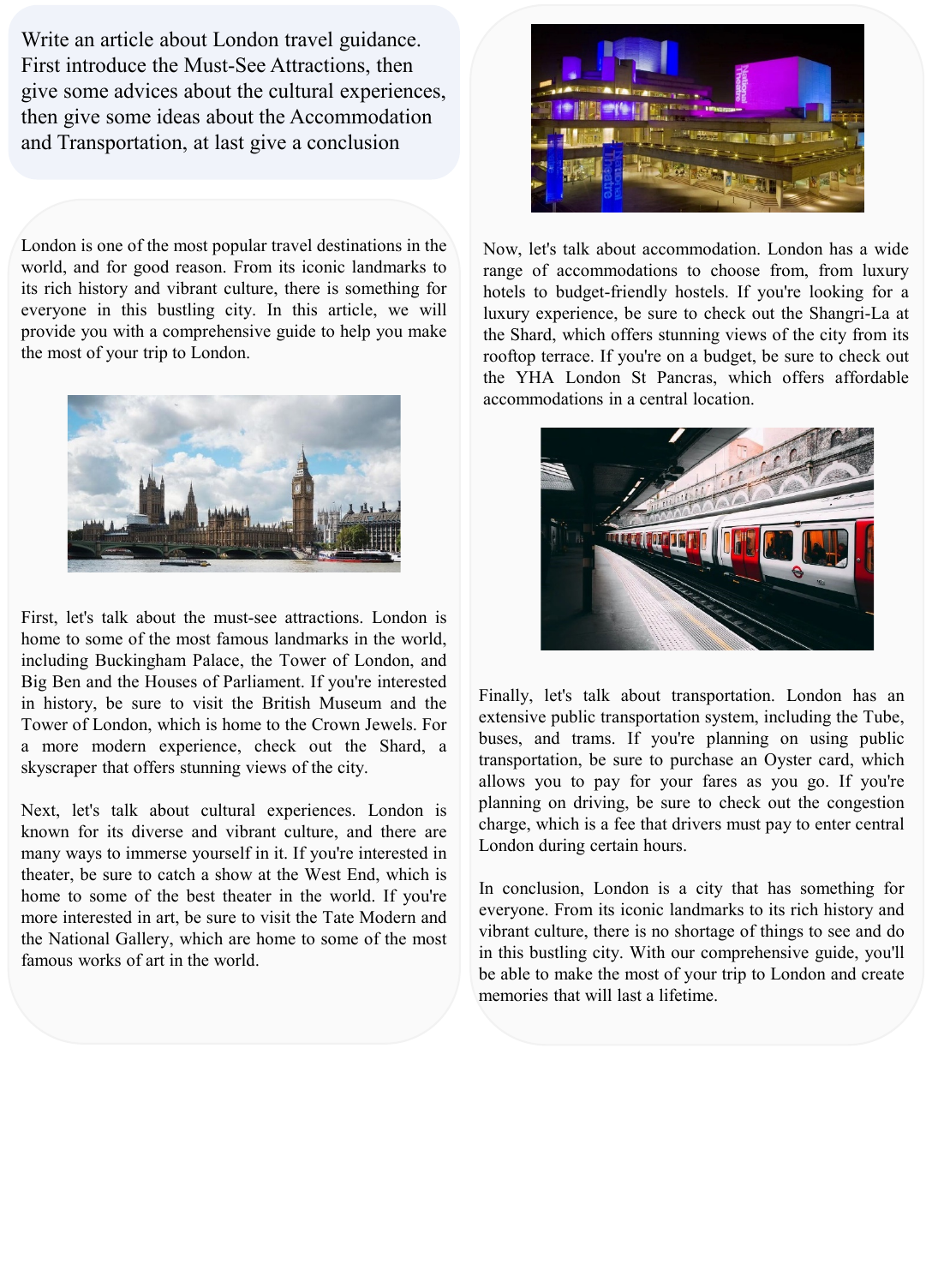}
    \vspace{-125pt}
    \caption{Interleaved composition generated by InternLM-XComposer.}
	\label{fig:article5}
\end{figure*}

\begin{figure*}[t]
	\centering
	\includegraphics[width=0.95\linewidth]{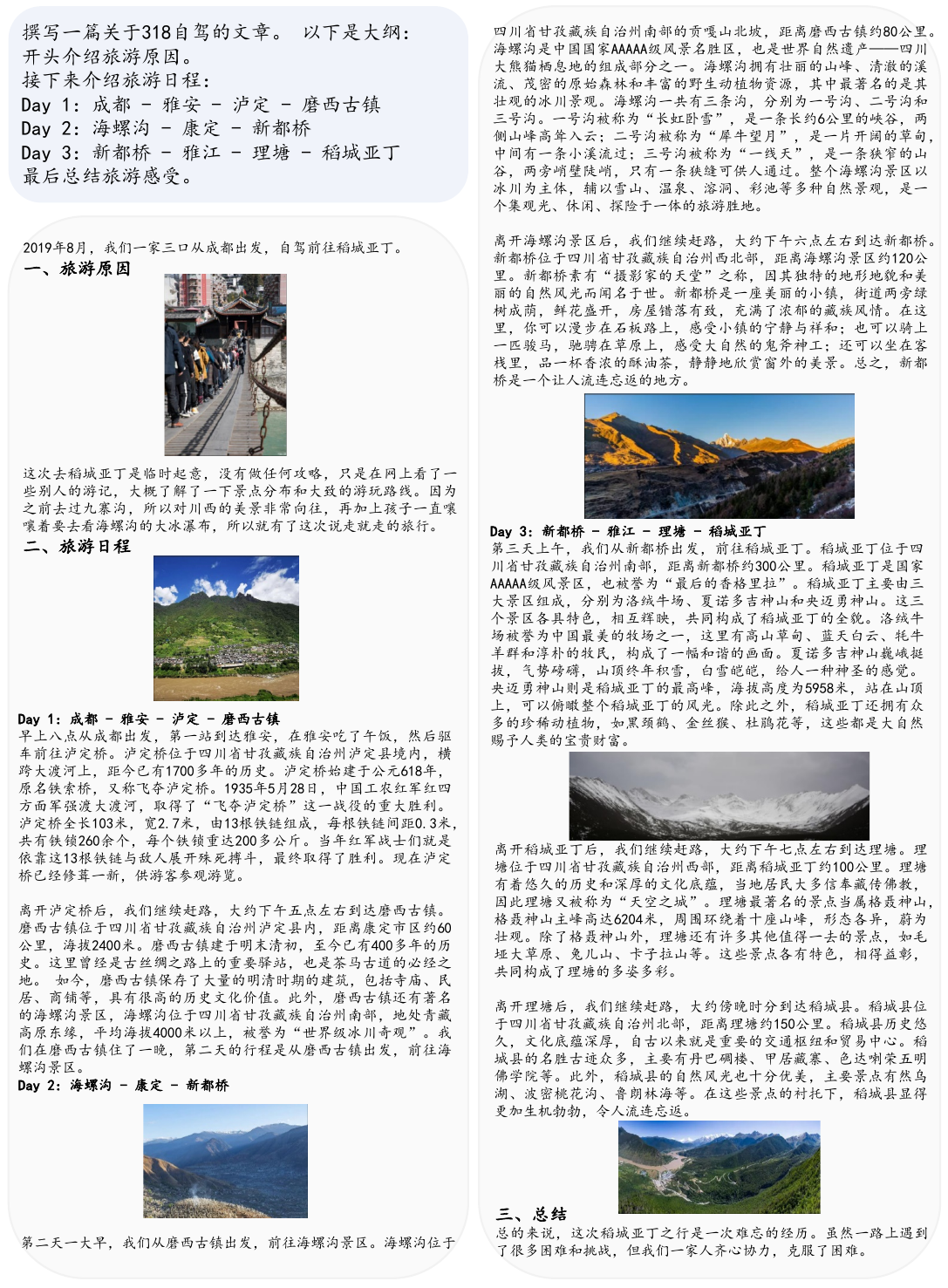}
    \vspace{-10pt}
    \caption{Interleaved composition generated by InternLM-XComposer.}
	\label{fig:article6}
\end{figure*}

\begin{figure*}[t!]
	\centering
	\includegraphics[width=0.95\linewidth]{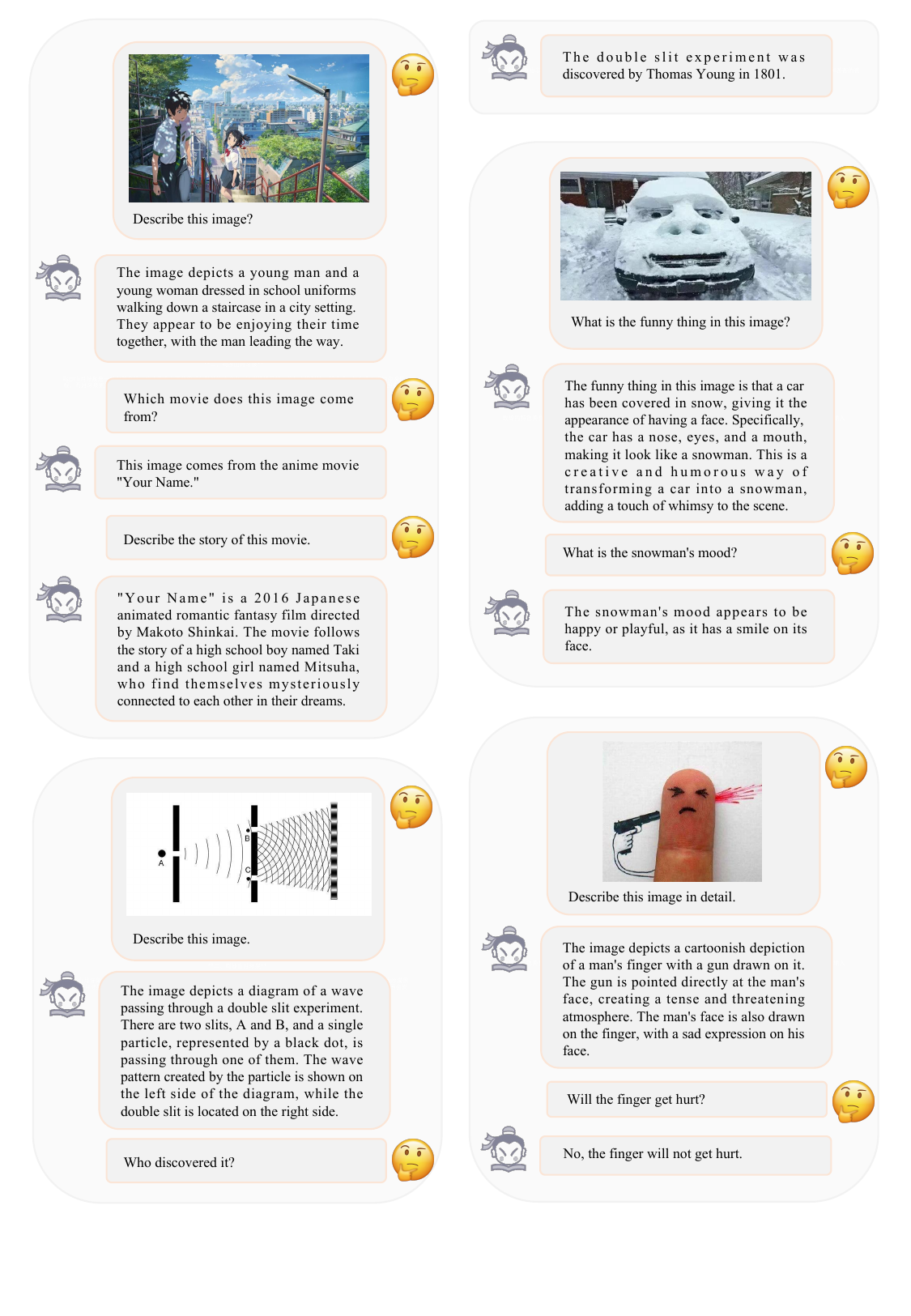}
    \vspace{-60pt}
	\caption{Chat with InternLM-XComposer.}
	\label{fig:conv1}
\end{figure*}

\begin{figure*}[t!]
	\centering
	\includegraphics[width=0.95\linewidth]{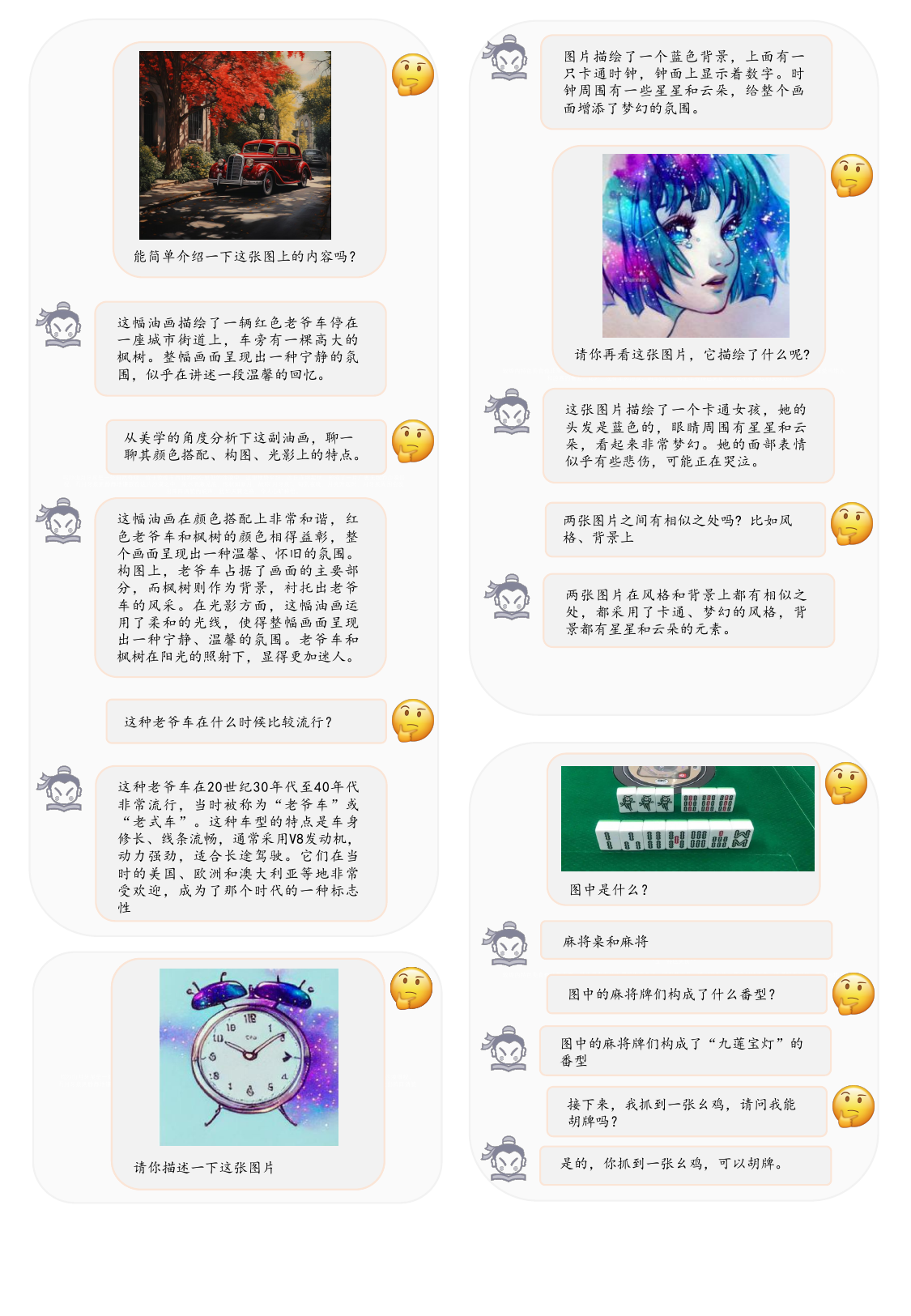}
    \vspace{-60pt}
	\caption{Chat with InternLM-XComposer. }
	\label{fig:conv2}
\end{figure*}

Figure~\ref{fig:article1}-\ref{fig:article6} shows some interleaved image-text compositions generated by InternLM-XComposer. 
Figure~\ref{fig:conv1} and Figure~\ref{fig:conv2} give some conversation cases when chatting with InternLM-XComposer. Our model exhibits an excellent ability in multilingual interleaved composition and conversation.

\clearpage

\end{document}